\documentclass[10pt,twocolumn,letterpaper]{article}

\usepackage{cvpr}              % To produce the CAMERA-READY version
\usepackage[accsupp]{axessibility}
\usepackage[dvipsnames]{xcolor}

\definecolor{cvprblue}{rgb}{0.21,0.49,0.74}
\usepackage[pagebackref,breaklinks,colorlinks,citecolor=cvprblue]{hyperref}
\usepackage{multirow}
\usepackage{graphicx}
\usepackage{comment}
\usepackage{makecell}
\usepackage{hyperref}

\def\paperID{12146} % *** Enter the Paper ID here
\def\confName{CVPR}
\def\confYear{2024}

\title{Benchmarking Segmentation Models with Mask-Preserved Attribute Editing}

\author{
Zijin Yin$^{1}$ \quad Kongming Liang$^{1 \ast}$ \quad Bing Li$^{2}$ \quad Zhanyu Ma$^{1}$ \quad Jun Guo $^{1}$\\
$^{1}$ Beijing University of Posts and Telecommunications \\
$^{2}$ King Abdullah University of Science and Technology\\
{\tt\small $^{1}$\{yinzijin2017, liangkongming, mazhanyu, guojun\}@bupt.edu.cn \quad $^{2}$bing.li@kaust.edu.sa}
}
\begin{document}
\twocolumn[{%
\renewcommand\twocolumn[1][]{#1}%
\maketitle
\vspace{-2.2em}
\begin{center}
    \centering
    \captionsetup{type=figure}
    \includegraphics[width=\textwidth]{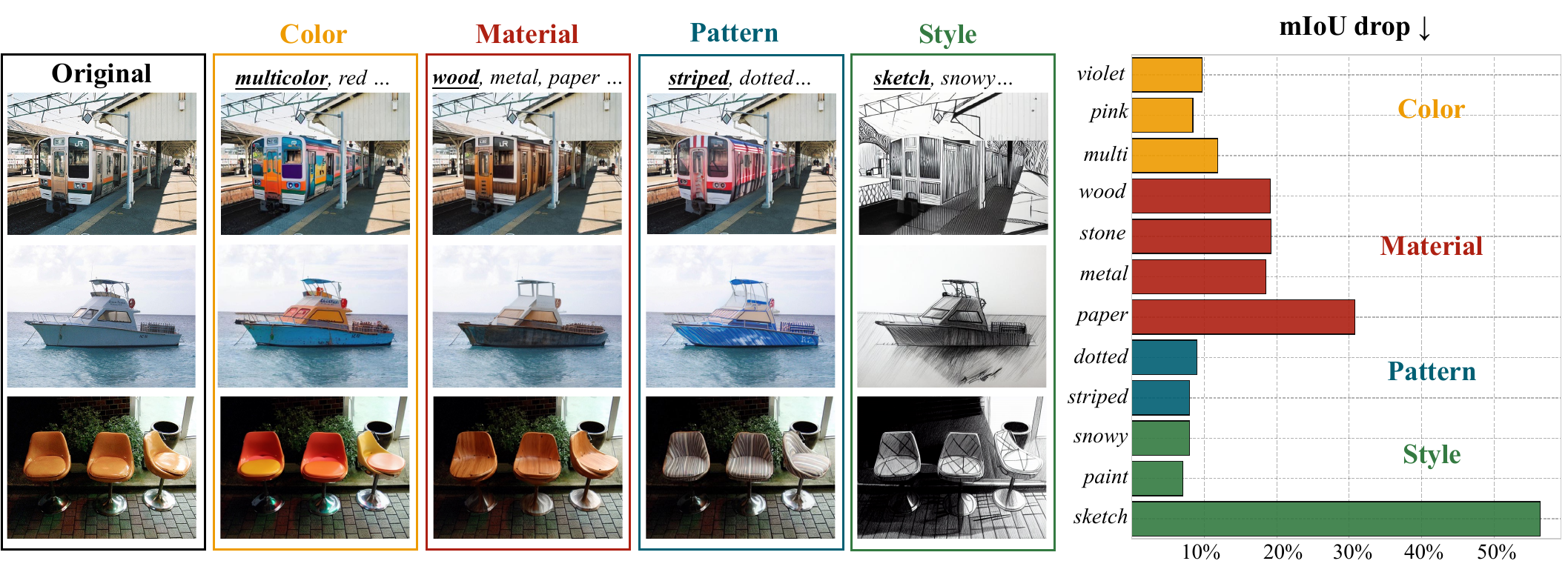}
    \vspace{-2.5em}
    \captionof{figure}{The illustration of the motivation of our work. \textbf{Left}: Our mask-preserved attribute editing pipeline generates testing images with various attribute changes for evaluating the robustness of segmentation methods to attribute variations. \textbf{Right:} The average performance drop of segmentation models on our generated data, shows the sensitivity to different types of attribute variations.}
    \label{fig:teaser}
\end{center}%
}]

\let\thefootnote\relax\footnotetext{$\ast$ Corresponding author.}

\begin{abstract}
\vspace{-1em}
When deploying segmentation models in practice, it is critical to evaluate their behaviors in varied and complex scenes.
Different from the previous evaluation paradigms only in consideration of global attribute variations (e.g. adverse weather), we investigate both local and global attribute variations for robustness evaluation. 
To achieve this, we construct a mask-preserved attribute editing pipeline to edit visual attributes of real images with precise control of structural information. Therefore, the original segmentation labels can be reused for the edited images.
Using our pipeline, we construct a benchmark covering both object and image attributes (e.g. color, material, pattern, style). We evaluate a broad variety of semantic segmentation models, spanning from conventional close-set models to recent open-vocabulary large models on their robustness to different types of variations. 
We find that both local and global attribute variations affect segmentation performances, and the sensitivity of models diverges across different variation types.
% Moreover, we find that stronger backbones and massive training data can not necessarily exhibit more robustness under our benchmark. 
We argue that local attributes have the same importance as global attributes, and should be considered in the robustness evaluation of segmentation models. Code: \url{https://github.com/PRIS-CV/Pascal-EA}.
\end{abstract}    
\vspace{-1em}
\section{Introduction}
\label{sec:intro}
In the last few years, deep-learning-based image segmentation models have witnessed remarkable progress in addressing complex visual scenes, enabling ubiquitous applications ranging from autonomous driving \cite{autonomous-driving} to medical image analysis \cite{medicalsam}. In order to ensure their reliability and robustness in real-world scenarios, it is desirable to evaluate them in varied and complex scenes in advance.

As to varied real-world scenarios, segmentation methods are required to be robust to various shifts in terms of local attribute variations (e.g. object color, material, pattern) and global attribute changes (e.g. image style). For example, a train can be painted with different colors and patterns, and a boat is composed of various materials (see Figure \ref{fig:teaser}). Moreover, segmentation methods may encounter images with global attribute changes such as variations in weather conditions and image styles in the real world. 
Nevertheless, the problem is: \textit{``How sensitive are existing segmentation models to local and global attribute changes?"} Regrettably, this research avenue remains under-explored.
The main challenge of robustness evaluation is the lack of high-quality test data with abundant local and global variations. In addition, even if the data can be collected, the annotation cost of the segmentation masks is quite high. 
The ACDC dataset \cite{acdc} manually collects samples with adverse weather in city streets. Sun \textit{et al.} \cite{shift} used simulation software to build a synthetic dataset, while Multi-weather Cityscapes \cite{multiweather} employed style-transfer models \cite{cyclegan} to generate data. However, they only evaluate segmentation models under global variations, such as weather and different scenes of city streets. Moreover, due to the limited capabilities of the simulation software itself and style-transfer models, their data are not consistent with realistic images.

In this paper, we provide a mask-preserved attribute editing pipeline for the robustness evaluation of segmentation models. The generated data can cover diverse variations including both object and image attributes. 
In particular, inspired by the impressive generation performance of text-guided image editing methods \cite{pnp, p2p, masactrl}, we leverage a pre-trained diffusion model and text instructions to transfer a testing image into images with various attributes via editing. 
To avoid manually annotating labels for edited images, we need to maintain the structural information of the image, such that the ground-truth segmentation mask of an original image is used as that of its edited ones. 
However, typical methods often improperly modify the structure of irrelevant regions (\eg removing an object from the background) during attribute editing. 
%This requires re-annotating the segmentation mask for the edited image since the ground-truth mask of the edited image is inconsistent with that of the original image. 
To address this issue, we propose mask-guided attention in the diffusion model \cite{ldm} to consistently edit target attributes at the object level and preserve the structure of an image. 
Thanks to the proposed module, our tool can edit images with attribute variations in a tuning-free manner, while avoiding manually annotating ground-truth segmentation masks.

Based on our mask-preserved attribute editing pipeline, we construct a benchmark covering various object and image attribute changes, and we evaluate the robustness of existing segmentation methods. We find that, as same as global attributes, local attribute variations also affect segmentation performances. And sensitivity of segmentation methods varies across different attributes. For example, performance declines most on object material variations (See Figure \ref{fig:teaser}). Moreover, the experimental results show open-vocabulary methods with stronger backbones and massive training data can not necessarily exhibit robustness, compared to conventional close-set methods. These findings suggest that object attribute variations have the same importance as image attribute variations to improve robustness.
%These findings suggest that improving robustness is a challenging problem and the object attributes should be taken into account.

%The test sets cover various attribute changes such as single attribute changes, combinations of multiple attribute changes, and different degrees of attribute changes. 
%We evaluate various segmentation methods on our generated test sets.
%The experimental results show the performance of both conventional close-set methods and recent open-vocabulary methods drop on our test sets, where we find these methods are sensitive to attribute changes. 
%Moreover, we also study the sensitivity of a segmentation method to specific attributes, where we find that the sensitivity of segmentation methods varies across different attributes. For example, the segmentation drops most on material attribute changes.

%\begin{itemize}
%\item  We benchmark various segmentation networks, which shows the performance of these networks drops on out-distribution data and in-distribution images containing various unseen object attributes. 
%\item We provide a pipeline that generates testing data controlled by text for segmentation methods. The pipeline  can be used to generate image object-level variation without introducing noticeable artifacts. Generate training images, and serve as data augmentation.  
%\item  We explore segmentation model sensitivity to local and global attributes such as  object attributes, image style,   and weather conditions.
%\end{itemize}
The main contributions are summarized as follows:
\begin{itemize}
    \item We provide a mask-preserved attribute editing pipeline that can change various attributes of real images without the requirement of re-collecting segmentation labels.
    \item We explore the robustness of existing segmentation models to both object and image attribute variations.
    %\item Our mask-preserved data generation pipeline generates high-quality test sets with various attribute changes without the need to annotate segmentation masks for generated images, given a segmentation dataset.  
    %We provide a tool that generates  a test set with various attribute changes in a zero-shot manner for benchmarking segmentation methods.
    % allows to consistently edit object-level and image-level attributes, which     
    \item We conduct extensive experiments and find that segmentation models exhibit varying sensitivity to different attribute variations. 
    %Data constructed by our pipeline have more reliability in evaluating segmentation than existing solutions.
    %find segmentation methods (including SAM [xxx]) are sensitive to both local and global attribute changes,  and the sensitivity of segmentation methods varies to different attributes.
\end{itemize}

\section{Related Work}
\noindent \textbf{Semantic Segmentation.} 
Conventional semantic segmentation methods \cite{deeplab, fastfcn, ocr} obtain significant performance across various benchmark datasets, but fail to generalize well on new environments. 
%Recently, many open-vocabulary segmentation frameworks \cite{cat, seem, ovseg, odise, x-decoder} are proposed by using language data as auxiliary supervision. 
Recently, since distinguished performance on alignment between visual and language data from visual language models (VLMs) \cite{clip, scaling}, many open-vocabulary segmentation frameworks \cite{seem, ovseg, odise} are proposed by using language data as auxiliary weak supervision. They normally generate class-agnostic masks, and then use the label embeddings from a Visual Language Model (VLM) to classify the proposal masks. 
%There are plenty of efforts to evaluate the robustness of segmentation models against domain shifts. 
Several benchmarks are proposed to evaluate segmentation models against domain shifts caused by attribute variations. 
The ACDC benchmark \cite{acdc} manually collects images and labels from multiple target domain sources.
SHIFT \cite{shift} employs simulation software to automatically generate data with continuous and discrete scene shifts in city streets. Other works \cite{adain, multiweather, foggydriving, robust, shift} synthesize images by training style transfer models \cite{cyclegan, stylegan}.
However, they only focus on global attribute variations, \textit{e.g.} weathers \cite{acdc} and scenes \cite{shift} in autonomous driving. 
Contrary to them, our research evaluates both local and global attribute variations which can depict real-world scenarios more comprehensively.

\noindent \textbf{Model Diagnosis.} 
Diagnosing the model's behavior and explaining its mistake is critical for understanding and building robustness. 
%Understanding a model’s behavior and explaining its mistakes is critical for building robust models.
In counterfactual explanation \cite{counterfactualreview} literature, it is common to explore what differences to input image will flip the prediction of the model.
%Originating from techniques used in tabular data \cite{tabular}, it is recently adopted to explain deep vision models behavior \cite{counterfactual,steex, beyond} by establishing the human-interpretable link between model predictions and perturbations to input images.
Previous methods explain decisions by mining similar failure samples from datasets to provide interpretations \cite{counterfactual, grounding, scout, adaptivetest}. However, the size of datasets constrains the variation types of similar images and hence hinders their overall interpretability. 
Recently, generation-based approaches \cite{texture-bias, discover-bugs, diffusion-explanation, khorram2022cycle, luo2023zero, zemni2023octet, li2023imagenet, lance, vendrow2023dataset, jeanneret2023text, howard2023probing} achieve remarkable progress in explaining model failures. 
Luo \textit{et al.} measure models' sensitivity to pre-defined attributes changes by optimization in latent space of StyleGAN \cite{stylegan}.
The closest works to ours are Prabhu \textit{et al.} \cite{lance} and ImageNet-E \cite{li2023imagenet}, which employ recent diffusion-based image editing techniques and achieve curated control on the diversity of image editing. 
However, their approaches inevitably interfere with irrelevant regions when editing object attributes and can not be applied to image segmentation tasks. 
Contrary to them, our proposed approach utilizes object masks to prevent potential interference and generate more realistic variations.

%\noindent \textbf{Text-Guided Image Editing.} 
%Recent GAN-based method \cite{vqgan} produce precise edits and diverse high-quality results, but require high computational cost. Diffusion models offer an efficient way to synthesize and edit images conditioned on text prompts. \cite{p2p, null, pnp, masactrl} edit both global and local aspects of the image by injecting the feature extracted from real images and controlling the attention maps in the forward diffusion process. 
%But their method inevitably interferes with other regions when local editing. 
%This issue is fatal in our setting, since segmentation methods may collapse under any unnecessary pixel perturbations. 
%Our proposed approach uses extra masks to prevent potential interference and generate more faithful image variations.

%-------------------------------------------------------------------------
\begin{figure*}[t]
    \centering
    \includegraphics[width=\textwidth]{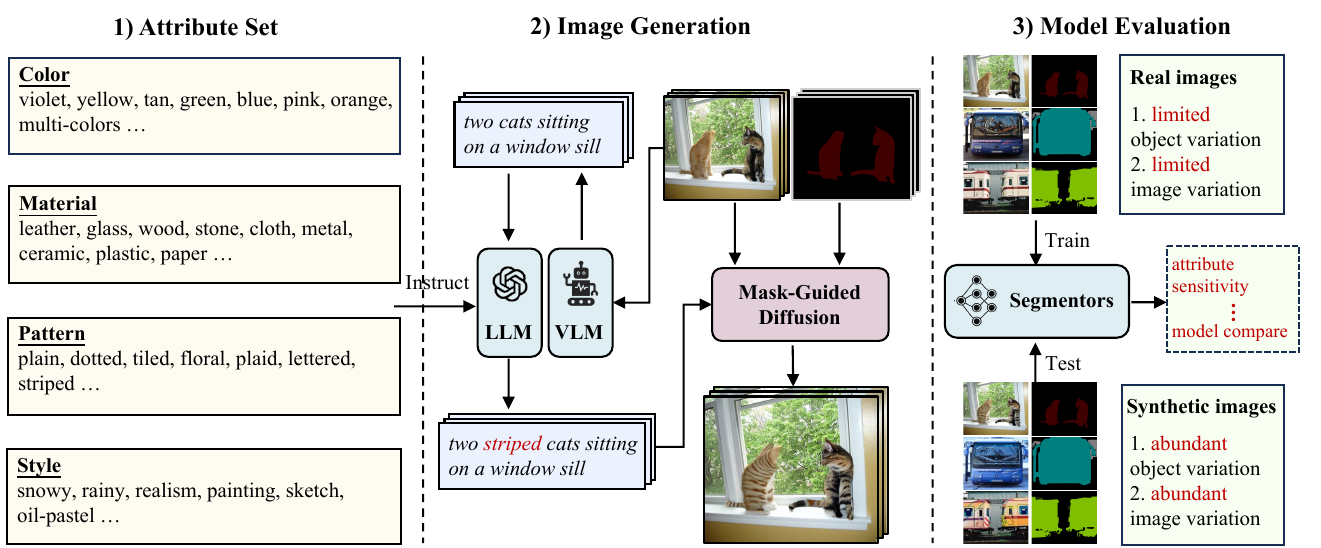}
    \caption{The illustration of our mask-preserved attribute editing pipeline. (1) We construct the Attribute Set which defines local and global variations. (2) We edit real images with different attribute variations with the collaboration of the large language model and diffusion model. (3) The robustness of segmentation models can be evaluated on the edited images against various types of attribute variations.}
    \label{fig:overview}
\end{figure*}

\section{Method}
\label{sec:method}
The overall pipeline of our method is illustrated in Figure \ref{fig:overview}. We first construct the Attribute Set which defines both local and global variations. Then we edit images with attribute variations via the collaboration of a large language model and diffusion model. Based on our design, we can generate samples with diverse objects and styles.
Finally, we evaluate the most representative segmentation models and investigate their robustness to different attribute variations. The details of each component are introduced as follows.

%-------------------------------------------------------------------------
\subsection{Attribute Set}
Generally, attributes~\cite{farhadi2009describing,liang-2019pami} are characteristic properties of objects and images, including color, shape, texture, and style.
A visual instance can be characterized by the composition of various attributes. For example, a cat may possess \textit{brown color, soft texture, dotted skin, feline shape}. In the existing visual datasets, some attributes, \textit{e.g.} the shape of cats, almost remain unchanged across diverse environments, while other attributes, \textit{e.g.} textures, materials, colors, have better diversity within a specific class \cite{net2vec}. 
To generate abundant evaluation samples, we construct an attribute set that encompasses all types of variations at both the object and image levels.

To trustfully evaluate the segmentation models, we investigate the attributes that tend to vary in the real world. In summary, we manually select attributes under four categories: (a) Local instance color, \textit{e.g.} blue, red. (b) Local instance material, \textit{e.g.} metal, wood, stone. (c) Local appearance pattern, \textit{e.g.} plain, dotted, lettered. (d) Global image style, \textit{e.g.} photo, painting, different weathers. Some examples are shown in Figure \ref{fig:overview}. 
By changing a single attribute and fixing the others, we can generate sufficient test samples with abundant variations.

 %-------------------------------------------------------------------------
\subsection{Image Generation}
\label{subsec:generation}
%With the collaboration of large language models \cite{llms} and Stable Diffusion \cite{ldm}
Since the generated images are used to evaluate segmentation models, they should have the following properties:
\begin{itemize}
    \item \textbf{Credibility}:  generated samples should be credible, adhering to real-world conditions, which implies that any modifications should be realistic.
    \item \textbf{Fidelity}: generated images should not disrupt irrelevant information and contravene layout defined by original semantic segmentation labels. (Since acquiring segmentation labels is laborious, we must ensure their correctness after generation.)
    
    %For example, we expect editing colors of the instance without affecting the background details and the original shape of the instance.
\end{itemize}
To ensure the above properties in our generated test samples, we design a mask-preserved attribute editing pipeline. 
The procedural details are illustrated as follows.

 %-------------------------------------------------------------------------
\noindent \textbf{Text Manipulation.} 
Since existing generative models are usually instructed by language, we first obtain the textual description of a real test image and incorporate different attributes linguistically. In this way, our method can be user-friendly. 
Specifically, we use a pre-trained large vision-language model (BLIP-2 \cite{blip2}) to acquire the textual description. 
To perform meaningful modification to text, we divided a text description into several editable components based on linguistic formal: domain, subjects and their adjectives (attributes), actions (verbs), and backgrounds (objects). For example, in \textit{``a photo of a white horse on the grass''}, the domain is \textit{``a photo''}, the subject is \textit{``horse''} and its adjective is \textit{``white''}, the action is \textit{``on''} and the background is \textit{``grass''}. 
Then we employ a language model (GPT-3.5 turbo) \cite{llms} to generate textual variations by instructions built using considered attributes. 
An example is \textit{change 'a horse on the grass' to 'a black horse on the grass'}. Please refer to our supplementary material for more prompts and examples.
Finally, we use the manipulated text descriptions as input prompts to guide image editing.

 %-------------------------------------------------------------------------
\noindent \textbf{Mask-Guided Diffusion.}
%We leverage state-of-the-art text-to-image algorithm Latent Diffusion Model \cite{ldm}, a.k.a Stable Diffusion (SD), as backbone. 
%The core component is denoising network U-Net \cite{unet} architecture, where the diffusion process is guided by the input text prompt via the attention mechanism.
There are several efforts \cite{sdedit, blended} to edit specific image areas by the guidance of object mask. However, most of them directly generate visual content according to the input text without injecting the features from the original images.
Therefore, they can only achieve limited performance in object attribute manipulation. 
Based on this observation, we utilize the setting suggested by \cite{p2p, pnp, masactrl, zero}, where features extracted from the real image are directly injected into the diffusion process of the generated image. 
However, we found that they can hardly change one specific portion of the original images without affecting others. For instance, manipulating the color of an object will marginally change the details of the adjacent background (refer to Figure \ref{fig:quality_assess}).  
This heavily diminishes the assessment reliability of generated test images, as the segmentation performance on whole images is susceptible to any pixel perturbation \cite{segattack}. 
From our perspective, the main reason is the attention maps in the diffusion process can only represent the rough spatial layout of objects but fail to accurately depict the detailed object localization. One trivial solution is directly replacing the background with pixel features from original images in each time step \cite{diffedit}. However, this induces a mass of artifacts in the object boundary.

% Similar investigations are also presented in \cite{densediffusion}, revealing spatial differences between attention maps and actual object localization in both cross-attention and self-attention layers. 

Inspired by the previous work~\cite{densediffusion}, we propose the Mask-Guided Attention which utilizes the object segmentation masks to rectify attention maps in the diffusion process. 
We denote a real image as $I \in \mathbb{R}^{3 \times H \times W}$ with height $H$ and width $W$ and its semantic layout label as $L$. Given the text guidance $P$, the objective of our method is to generate image $I^{*}$ which complies with $P$, and strictly preserves the semantic layout defined by $L$. Concretely, supposed that a binary object mask $S \in \mathbb{R}^{H \times W}$ is extracted from the semantic map $L$, we rectify the attention map as follows:

\begin{figure}
    \centering
    \includegraphics[width=\columnwidth]{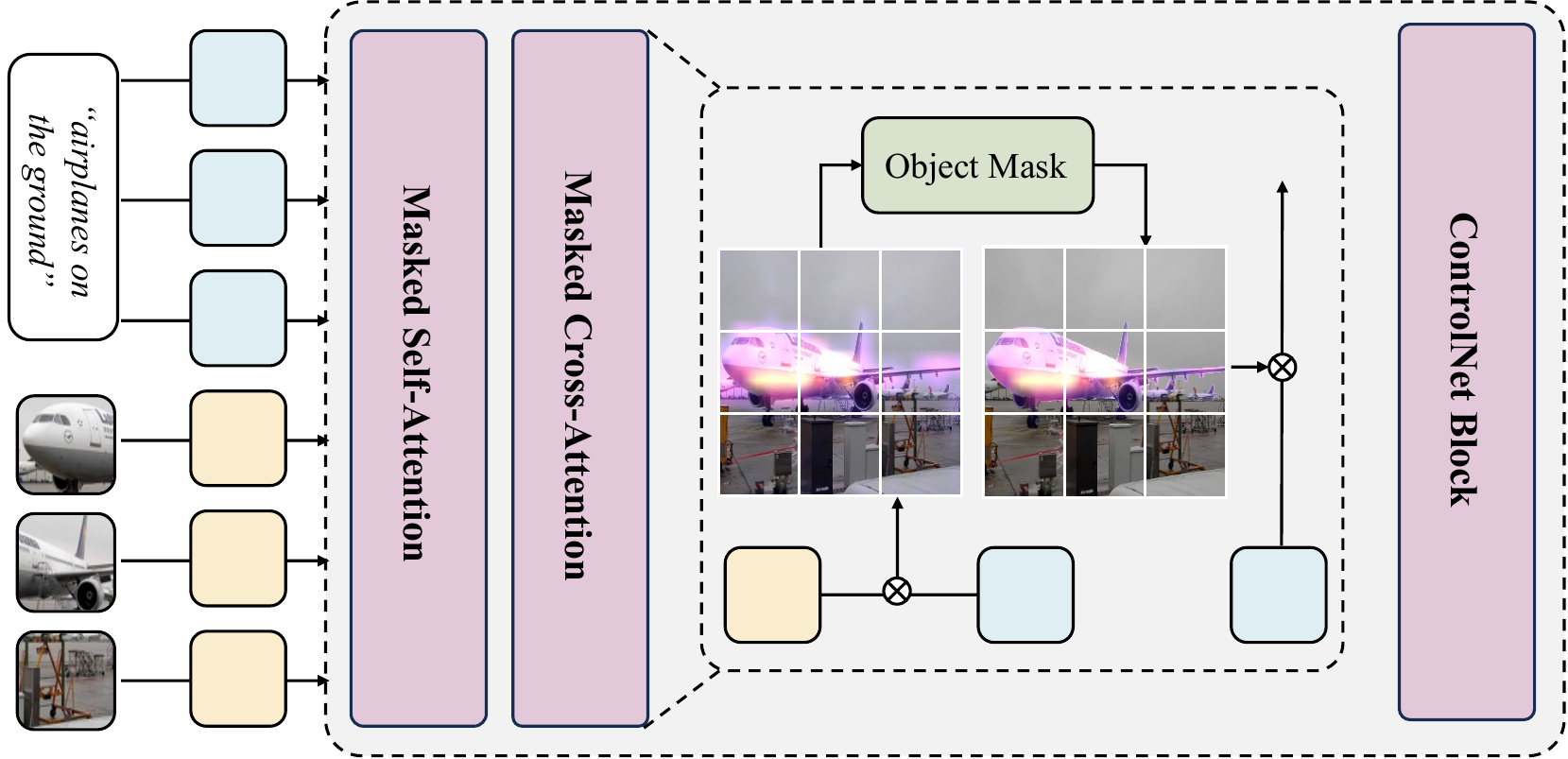}
    \caption{The illustration of block in our method. Our attention mechanism utilizes an object mask to rectify attention maps in self-attention and cross-attention layers. We adopt ControlNet block \cite{controlnet} to further restrict the semantic layout.}
    \vspace{-1em}
    \label{fig:attention}
\end{figure}

\begin{equation}
    A = \operatorname{softmax}\left(\frac{Q K^{\top}+R \odot \phi(S^{'})}{\sqrt{d}}\right),
\end{equation}
where the query-key pair conditioning map $R \in \mathbb{R}^{\mid Q|\times| K \mid}$ defines whether to rectify the attention score for a particular pair, we set different configurations of $R$ in self and cross attention layers. $\phi(\cdot)$ is a flatten function, and $S^{'}$ is a rectification map confining the region of attention scores which is defined as:
\begin{equation}
    S^{'}(x, y)= \begin{cases}0 &  S(x, y)=1 \\ -\infty & \text { otherwise. }
    \end{cases}
\end{equation}

In the cross-attention layer, the semantic knowledge of textual features is incorporated into intermediate visual features in which objects appearance and image layouts gradually emerge. The attention maps between each text token and visual token determine the intensity and area of influence of corresponding text words. We rectify the corresponding region and intensity of cross-attention scores of text words, the conditioning map $R$ at location $(i,j)$ is
\begin{equation}
\label{equ:degree}
    R(i,j) = \begin{cases} 
    c & i \in \{K\}, \forall j \\ 
    0 & \text{otherwise},   
    \end{cases}
\end{equation}
where ${K}$ is the set of token indexes corresponding to the object mask $S$ and $c$ indicates the strengthens or weakens the extent to which text tokens affect the resulting image.

In the self-attention layer, the intermediate visual features compute affinities with each other, allowing to retain the coherent spatial layout and shape details. We leverage the object mask $S$ to constrain interactions between different regions. For example, we retain the attention scores between tokens within the same regions and suppress that between tokens within different regions. Concretely, we define the conditioning map $R$ at location $(i,j)$ as:
\begin{equation}
    R(i,j) = \begin{cases}
        1 & \phi(S)(i)=1 \text { and } \phi(S)(j)=1 \\
        0 & \text{otherwise}.
    \end{cases}
\end{equation}

\noindent We incorporate the designed attention mechanism into decoding blocks of U-Net \cite{unet} in the Latent Diffusion Model \cite{ldm}. In global style editing, we denote $S$ as a matrix of ones where every entry is equal to one.

In order to further control that generated images rigorously adhere to their original semantic layout, we integrate existing controllable generative module ControlNet \cite{controlnet} into each blocks. 
The whole designed block is illustrated in Figure \ref{fig:attention}. 
In this way, our pipeline can obtain samples with abundant diversity, without changing their original semantic segmentation labels. And we can hence expand existing benchmarks requiring no laborious label collection.

\begin{table*}[t]
\renewcommand\arraystretch{1.05}
\centering
\caption{Quantitative results of all segmentation models under different attribute variations in our Pascal-EA. The results are reported using mIoU ($\uparrow$), and the best results of close-set methods and open-vocabulary methods are separately bold. We replace the original validation set with the reconstructed images obtained by our pipeline (denoted as Val).} 
\label{table:main}
\resizebox{\linewidth}{!}{
\begin{tabular}{cc c ccc cccc cc ccc c}
\toprule
\multirow{2}{*}{Method} & \multirow{2}{*}{Backbone} & \multirow{2}{*}{\textbf{Val}} & \multicolumn{3}{c}{\textbf{Color}}  & \multicolumn{4}{c}{\textbf{Material}} & \multicolumn{2}{c}{\textbf{Pattern}}   & \multicolumn{3}{c}{\textbf{Style}} & \multirow{2}{*}{\textbf{Overall}} \\
  \cmidrule(lr){4-6}  \cmidrule(lr){7-10}  \cmidrule(lr){11-12} \cmidrule(lr){13-15}
 &        &      & violet & pink & multi-color & wood & stone & metal & paper & dotted & striped & snowy & painting & sketch &         \\
\midrule
DeepLabV3+ \cite{deeplabv3plus}& RN-50      & 73.65  &  66.46  & 67.29 & 63.34 & 52.96  & 54.10 & 55.07 & 41.98 & 62.85  & 63.59   & 65.14 &  61.39  & 15.33  & 55.79 ({\textcolor{red}{-17.86}})       \\
OCRNet \cite{ocr}  & HR-48   & 75.32  &  67.78  & 69.98 & 65.65 & 56.79  & 56.76 & 58.69 & 46.42 & 65.41  & 67.41   & 67.71 &  65.45  & 22.06  & 59.18 (\textcolor{red}{-16.14})   \\
Segmenter \cite{segmenter} &  ViT-B    &  80.66 &  74.14  & 75.2 & 72.77 & \textbf{65.84} & \textbf{63.11} & \textbf{67.15} & \textbf{54.89} & 74.48  & 75.26 & 75.22 & 75.86  & \textbf{31.19} &    \textbf{67.09} (\textbf{\textcolor{red}{-13.57}})  \\
Segformer \cite{segformer} &  MIT-B3    & 79.45 & 71.06	& 72.37	& 69.01	& 60.55	& 61.6	& 63.02	& 51.88	& 70.8	& 72.02	& 71.91	& 71.23	& 21.45  &  63.08(\textcolor{red}{-16.38})     \\
%UPerNet \cite{convnext}  & ConvNeXt-B      &   &    &  &  &  &  &  &  &   &   &  &   &  &      \\
Mask2Former \cite{mask2former} & Swin-B    & \textbf{90.81}  &  \textbf{80.10}  & \textbf{80.15} & \textbf{72.92} & 61.14  & 62.92 & 62.77 & 49.48 & \textbf{77.29}  & \textbf{77.78}  & \textbf{76.21} &  \textbf{77.06}  & 21.82  & 66.64 (\textcolor{red}{-24.17})     \\
\midrule
CATSeg \cite{cat}             & Swin-B    & \textbf{95.59}  &  \textbf{88.06}  & \textbf{88.51} & \textbf{87.28} & \textbf{82.93}  & \textbf{81.96} & \textbf{82.26} & \textbf{74.79} & \textbf{88.63}  & \textbf{90.30}   & \textbf{90.06} &  \textbf{92.95}  & \textbf{48.06}  & \textbf{82.98} (\textbf{\textcolor{red}{-12.61}})   \\
OVSeg \cite{ovseg}             & Swin-B    & 92.83  &  83.23  & 85.84 & 85.17 & 79.08  & 77.01 & 79.70 & 70.93 & 86.12  & 86.06   & 86.93 &  89.74  & 42.79  & 79.38 (\textcolor{red}{-13.45}) \\
ODISE \cite{odise}             & SD-1.5   & 91.54  &  71.96  & 72.05 & 70.05 & 70.23  & 73.46 & 64.50 & 54.59 & 81.55  & 84.33   & 82.61 &  85.97  & 26.32  & 69.79 (\textcolor{red}{-21.75})   \\
X-Decoder \cite{x-decoder}     & Focal-T  & 90.42  &  81.70  & 83.89 & 79.36 & 70.59  & 69.64 & 72.72 & 52.28 & 83.58  & 83.85   & 83.22 &  87.05  & 34.99  & 73.57 (\textcolor{red}{-16.85})   \\
SEEM \cite{seem}               & Focal-T  &  89.21	& 80.30	& 82.74	& 78.55	& 70.79	& 69.56	& 71.37	& 56.94	& 81.76	& 82.42	& 83.71 & 	84.91	& 34.45	& 73.13 (\textcolor{red}{-16.09})    \\
SEEM \cite{seem}               & SAM-B    & 89.10  &  77.14  & 79.48 & 72.20 & 61.41  & 63.37 & 64.98 & 50.95 & 77.47  & 77.80   & 76.21 &  77.06  & 28.18  & 67.19 (\textcolor{red}{-21.91})   \\
\midrule
\multicolumn{2}{c}{\multirow{2}{*}{\textbf{Overall}}}   &  \multirow{2}{*}{86.21} &  76.54  & 77.95  & 74.21 & 66.57 & 66.68 & 67.48 & 55.00 & 77.27 & 78.26 & 78.08 & 78.97 & 29.69  &    \\
& & & \textcolor{red}{-9.68} & \textcolor{red}{-8.27} & \textcolor{red}{-12.01} & \textcolor{red}{-19.65} & \textcolor{red}{-19.54} & \textcolor{red}{-18.74} & \textcolor{red}{-31.22} & \textcolor{red}{-8.95} & \textcolor{red}{-7.96} & \textcolor{red}{-8.14} & \textcolor{red}{-7.25} & \textcolor{red}{-56.53} & \\
\bottomrule
\end{tabular}}
\end{table*}

\section{Experiments}
\label{sec:experiment}
In this section, we first introduce the experimental setups. Then we benchmark various segmentation models on our edited images and discuss on the results. At last, we assess the quality of our edited images. 
%Next, we evaluate diverse segmentation models on our generated benchmark, compare their behaviors, and investigate the effects of different variations. 

\subsection{Experimental Setup}
\noindent \textbf{Benchmark.} 
We construct our benchmark \textbf{Pascal-EA(Editable Attributes)} by editing images in the validation set of Pascal VOC dataset \cite{pascal}.
Most samples in Pascal VOC \cite{pascal} are object-centric, making it more suitable for exploring the impact of local variables on segmentation performance. 
Considering potential errors in generated images, we manually discarded ones with noticeable errors to ensure the quality of the benchmark. 
Since we want to stress-test models, we manually choose the following uncommon attributes to construct the benchmark. 
(1) \textit{object colors}: violet, pink, and multicolor, which provide a deviation from objects' common colors. 
(2) \textit{object material}: wood, stone, metal and paper.
(3) \textit{object pattern}: dotted and striped, which have unusual appearance compared to common objects.
(4) \textit{image style}: snow, painting, and sketch which provides blurring circumstances, abstract descriptions of realistic objects, and absence of color and texture visual cues, respectively. Please refer to our supplementary material for qualitative results.

\noindent \textbf{Target Models.}  
We conduct evaluation experiments on various architectures, including close-set and open-vocabulary methods. 
For close-set methods, we adopt convolution-based models DeepLabV3+ \cite{deeplabv3plus}, OCRNet \cite{ocr}, and transformer-based models Segmenter \cite{segmenter}, Segformer \cite{segformer}, and the universal image segmentation architecture Mask2Former\cite{mask2former}. 
For open-vocabulary methods, we adopt transformer-based approaches CATSeg \cite{cat} and OVSeg \cite{ovseg}, the typical diffusion-based method ODISE \cite{odise}, and two SOTA generalized frameworks X-Decoder \cite{x-decoder} and SEEM \cite{seem}. We implement target models with MMsegmentation \cite{mmseg} and Detectron2 \cite{detectron2}. All models are trained in the training set of Pascal VOC \cite{pascal}, and we use the released official weights and original recipes to compare at their best.

\noindent \textbf{Metric.} We report the results of mIoU as the metric.
 
\begin{table*}[t]
\centering
\caption{ mIoU ($\uparrow$) of four augmentation algorithms on our Pascal-EA. The best results are in bold.}
\label{table:augmentation}
\resizebox{\linewidth}{!}{
\begin{tabular}{c c ccc cccc cc ccc}
\toprule
\multirow{2}{*}{Method}  & \multirow{2}{*}{\textbf{Val}} & \multicolumn{3}{c}{\textbf{Color}}  & \multicolumn{4}{c}{\textbf{Material}} & \multicolumn{2}{c}{\textbf{Pattern}}   & \multicolumn{3}{c}{\textbf{Style}} \\
  \cmidrule(lr){3-5}  \cmidrule(lr){6-9}  \cmidrule(lr){10-11} \cmidrule(lr){12-14}
 &   & violet & pink & multi-color & wood & stone & metal & paper & dotted & striped & snowy & painting & sketch  \\
\midrule
Mask2Former \cite{mask2former}& 90.81  &  80.10  & 80.15 & 72.92 & 61.14  & 62.92 & 62.77 & 49.48 & 77.29  & 77.78  & 76.21 &  77.06  & 21.82      \\
\midrule
CutOut \cite{cutout}   & 91.44  &  80.68 & 82.02 & 73.44 & 62.23  & 64.70 & 60.55 & \textbf{50.28} &  75.32  & 78.32  & 75.95 &  75.41  &  19.55   \\
CutMix \cite{cutmix}   & \textbf{92.01} & \textbf{80.79}  & \textbf{83.45} & \textbf{75.53} &  \textbf{64.81} & \textbf{66.68} & \textbf{62.87} & 49.83 & \textbf{77.84}  & \textbf{78.73} & 76.03 & 76.48 &  21.62    \\
Digital Corruption \cite{weather}  & 91.19 & 73.29  & 75.47 & 69.63 &  57.96 & 55.51 & 56.40 & 45.01 &  75.16 & 75.79  & 79.22 &  77.98  & 22.59        \\
AugMix \cite{augmix}   & 91.88 & 75.38  & 76.49 & 70.44 & 60.29 & 58.92 & 58.73 & 48.84 &  75.33 &  76.01 & \textbf{80.46}  &  \textbf{80.92}  &  \textbf{23.05}      \\
\bottomrule
\end{tabular}}
\vspace{-0.5em}
\end{table*}

\subsection{Evaluation Results} 
%We benchmark various segmentation models in our generated image and investigate the impacts of different variations.
Table \ref{table:main} presents segmentation performances under different attribute variations.
%To explore the effects of different types of variations, we calculate the average decline in performance across all models for each type and present results in the final row. 
Initially, our generated images, featuring unusual visual cues, reduce performance across all models to varying extents.
However, it is noteworthy that the models exhibit a greater vulnerability to material alteration than to adjustments in other local attributes, such as color and pattern. 
For instance, under the material-based variations, performance decline ranged from 18.74\% to 31.22\%, whereas under the color-based variations, the decline ranged from 8.27\% to 12.01\%. Changes in the object's appearance pattern have less impact on the models' performance compared to color and material.
Since the choice of material significantly influences the visual texture of an object's appearance, we speculate that segmentation models are significantly sensitive to texture, which aligns with observations found in classifiers \cite{texture-bias}.
Furthermore, results reveal a notable disparity in model performance when comparing sketch images to other global styles. For instance, performances decrease 56.53\% under sketch style variation, while dropping 7.25\% and 8.14\%  under painting and snow scenarios.
We argue that sketch images introduce substantial perturbations to model performance as they represent visual content solely through line-based representations without any texture and color information. 
%In summary, we can propose that segmentation models rely not only on shape cues but also on other visual factors such as texture and color information.

To compare the overall robustness of models, we also compute the average performance decline under all attributes in the last column of Table \ref{table:main}. 
Firstly, in close-set methods, we observe that Segmenter \cite{segmenter} exhibits best performances under material variations while Mask2Former \cite{mask2former} performs best in others. And we notice that the overall performance decline of DeepLabV3+ \cite{deeplabv3plus} and OCRNet \cite{ocr} is even less than that observed in transformer architectures such as Segformer \cite{segformer} and Mask2former \cite{mask2former}.
This implies that recent transformer-based models have greater segmentation accuracy than CNN-based methods, but they do not necessarily show an improvement in robustness. Previous studies \cite{de2023reliability} have shown that recent transformers are remarkably more robust than baseline models as the domain gap is large, which is opposite to our findings. We think the reason may be our edited images have a smaller domain gap compared to \cite{de2023reliability}.
Secondly, in open-vocabulary frameworks, we notice ODISE \cite{odise} exhibits considerably inferior performance compared to CATSeg \cite{cat} and OVSeg \cite{ovseg}. Since they have only a discrepancy in mask proposal backbone, we argue that Stable Diffusion \cite{ldm} possesses comparatively worse robustness than specialized backbone \cite{swin} in segmentation. 
Besides, two holistic-trained frameworks X-Decoder \cite{x-decoder} and SEEM \cite{seem} are worse than other approaches. 
We speculate the reason relies on the utilization of the Vision-Language Model (VLM) which classifies proposed masks. 
These two frameworks retrain a vision-language alignment space from scratch, and other approaches leverage off-the-shelf CLIP \cite{clip} trained with much more abundant data. 
Finally, we also evaluate the performance of the recent Segment-Any-Thing (SAM) \cite{sam} backbone using the SEEM \cite{seem} framework. SEEM method with different backbone exhibits similar performances in the in-distribution set while gaining considerable disparity in our generated benchmarks, in which the SAM fails to show better robustness.
In summary, all phenomenons reveal that better in-distribution performances do not indicate better performance under our variations, and stronger backbones and more training data do not necessarily bring robustness improvement. Please refer to our supplementary materials for more results and analysis.

\subsection{Discussion}
\noindent \textbf{Effectiveness of augmentation techniques.}
We also conduct experiments to explore the effectiveness of existing data augmentation algorithms. We select several representative methods: CutOut \cite{cutout}, CutMix \cite{cutmix}, AugMix \cite{augmix} and Digital Corruption \cite{weather}, and use vanilla Mask2Former \cite{mask2former} as a baseline model. 
The results are presented in Table \ref{table:augmentation}. We can obtain several observations: (1) All augmentation methods improve in-distribution performances. (2) While CutOut \cite{cutout} and CutMix \cite{cutmix} exhibit more robustness on object-level variations of color, material, and pattern, they can not improve performances on image-level style variations. Compared to the previous two, Digital Corruption \cite{weather} and AugMix \cite{augmix} exhibit opposite characteristics.
We speculate that the reason behind these divergent behaviors is local object variation edits the visual content information whereas global variation tends to edit visual styles. Since Digital Corruption \cite{weather} and AugMix \cite{augmix} change the style information of images by methods such as brightness and sharpness, the model trained on it can only exhibit advancement on global style variations. However, CutOut \cite{cutout} and CutMix \cite{cutmix} directly change the visual content of specific regions, they can act better under object variations. 
%In summary, these methods are unable to solve both object-level and image-level variations. This may serve as evidence that our benchmark is meaningful for comprehensive evaluations.

\begin{figure}
    \centering
    \includegraphics[width=\columnwidth]{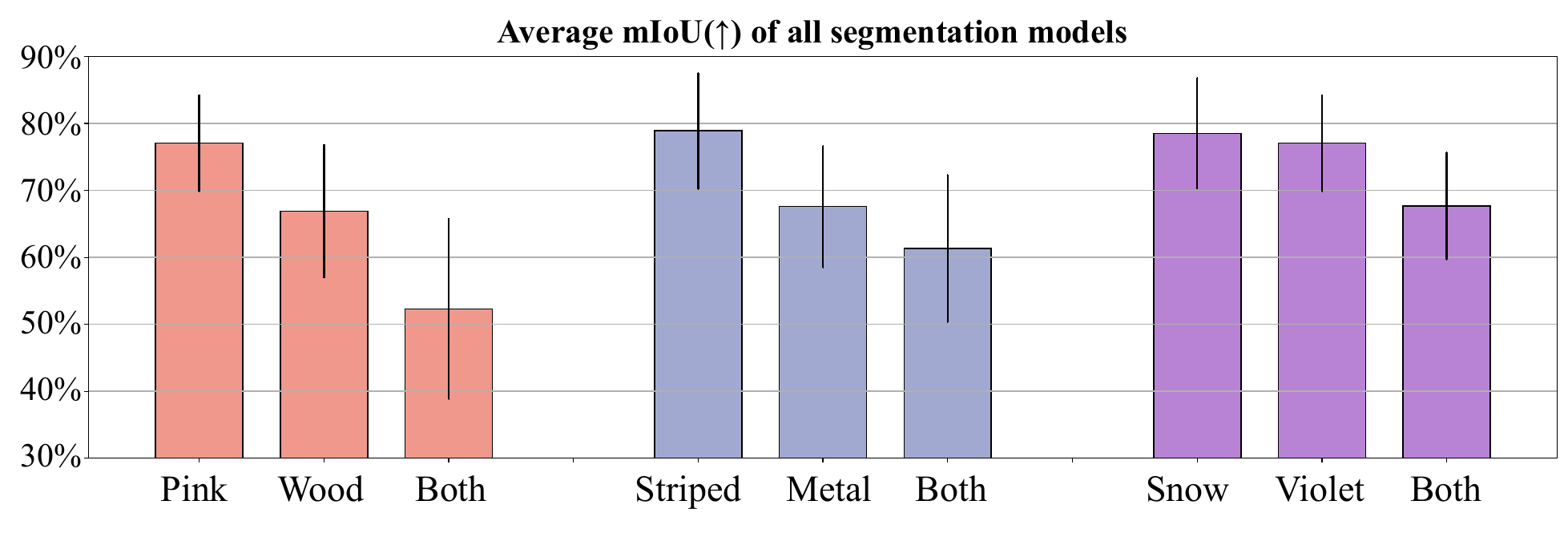}
    \vspace{-1.5em}
    \caption{Average mIoU ($\uparrow$) of all segmentation methods under the combination of two different attribute variations.}
    \vspace{-1em}
    \label{fig:multi-attributes}
\end{figure}

\noindent \textbf{Multiple Attribute Variation.} 
We conduct a statistical analysis of the average model performances when exposed to samples with a combination of two distinct attribute variations. The results are presented in Figure \ref{fig:multi-attributes}. It is evident that when two types of variation are combined, the overall performance gains more deterioration compared to that under a single one. 
This observation indicates that addressing multiple attribute variations poses a greater challenge than dealing with a single in isolation. 

\noindent \textbf{Variation Degree.} 
The extent to which edited images deviated from the original ones can be adjusted by coefficient $c$ in the Equation \ref{equ:degree}, higher value of $c$ means greater variation. 
We adjust different values of $c$ in the image editing process to investigate the impact of variation degree on segmentation performance. 
Experimental results are presented in Table \ref{table:variation_degree}. 
We notice a gradual degradation in all models' performances as the degree of variation intensifies across all scenarios. 
This reveals our images do not make models collapse like the adversarial attack method \cite{segattack}, and hence have the capability of continuous stress-testing models.
\begin{table}[t]
\renewcommand\arraystretch{1.05}
\centering
\caption{mIoU ($\uparrow$) of five segmentation methods under three adverse weather conditions with different variation degrees. Segmentation performances gradually deteriorate as shift increases.} 
\label{table:variation_degree}
\resizebox{\columnwidth}{!}{
\begin{tabular}{cc cc cc cc}
\toprule
 & \multirow{2}{*}{\textbf{Val}} & \multicolumn{3}{c}{\textbf{Style:snow}} & \multicolumn{3}{c}{\textbf{Material: Wood}} \\
\cmidrule(lr){3-5} \cmidrule(lr){6-8} 
                  &       & $c=1.0$         & $c=2.0$        & $c=3.0$         & $c=1.0$        & $c=2.0$        & $c=3.0$        \\
\midrule
DeepLabV3+\cite{deeplab}          & 73.65  &  65.14   & 63.79    &  59.20  &  52.96  & 49.38   &  45.82   \\    
OCRNet \cite{ocr}                 & 75.32  &  67.71   & 63.55    &  60.29  &  56.79  & 51.67   &  46.23   \\
Segmenter \cite{segmenter}        & 80.66  &  75.22   & 68.03    &  62.09  &  \textbf{65.84}  & \textbf{60.28}   &  \textbf{56.89}   \\
Segformer \cite{segformer}        & 79.45  &  71.91   & 67.85    &  60.88  &  60.55  & 55.79   &  49.75   \\
Mask2Former \cite{mask2former}    & \textbf{90.81}  &  \textbf{76.21}   & \textbf{70.24}    &  \textbf{63.49}  &  61.14  & 57.34   &  51.63   \\
\bottomrule
\end{tabular}}
\vspace{-1.3em}
\end{table}

\noindent \textbf{Practical application.} 
Our pipeline could also be utilized as a data augmentation approach to improve robustness of segmentation methods under hard samples. Please refer to our supplementary material for results details.

%\begin{table}[]
%\caption{Summary of existing benchmarks for semantic segmentation robustness evaluation.}
%\label{table:benchmark}
%\resizebox{\columnwidth}{!}{
%\begin{tabular}{lccc}
%\toprule
%Dataset                             & Source             & Scene          & Variation \\
%\midrule
%ACDC \cite{acdc}                    & real               & cityscape      & weather, daytime          \\
%Dark Zurich \cite{darkzurich}       & real               & cityscape      & daytime     \\
%MWC \cite{multiweather}             & style transfer     & cityscape      & weather        \\
%Fog Cityscapes \cite{foggydriving}     & simulation         & cityscape      & fogging     \\
%SHIFT \cite{shift}                  & software           & cityscape      & weather, daytime    \\
%Stylized COCO \cite{trapped}        & style transfer     & general        & object texture \\
%BigDatasetGAN \cite{} & synthetic & universal & - & \ding{55}     \\
%DiffuMask \cite{diffumask} & synthetic & universal & - &  \ding{55}    \\
%\midrule
%\multirow{2}{*}{\textbf{Ours}}      & \multirow{2}{*}{diffusion} & \multirow{2}{*}{general} &  object attributes   \\
% & &  &  image styles    \\
%\bottomrule
%\end{tabular}}
%\end{table}

\subsection{Image Quality Assessment}
In this section, we assess the quality of our generated images by comparing them with previous diffusion-based image editing approaches, and popular benchmarks of segmentation robustness evaluation.

\noindent \textbf{Comparison with diffusion-based image editing methods.}
\label{subsec:compare_diffusion}
Since our proposed pipeline serves as an evaluator for segmentation models, the reliability of edited images is our primary concern. We focus on object internal structure maintenance and no disruption to irrelevant information. Thus, we adopt LPIPS and DINO Dist \cite{dinodist}, which measure structural distances, as our main metrics.
We compare our proposed method with state-of-the-art image editing algorithms PnP \cite{pnp}, Prompt-to-Prompt \cite{p2p}. Since there are several efforts to utilize object masks as guidance, such as DiffEdit \cite{diffedit} and Blended Latent Diffusion \cite{blended}, we take them as additional comparisons. We also conducted an ablation study to demonstrate the effectiveness of the Mask-Guided Attention mechanism and ControlNet. 
Evaluations are performed on tasks of random replacing object color and material using Pascal VOC dataset \cite{pascal}. 

\begin{figure*}[t]
    \centering
    \includegraphics[width=\textwidth]{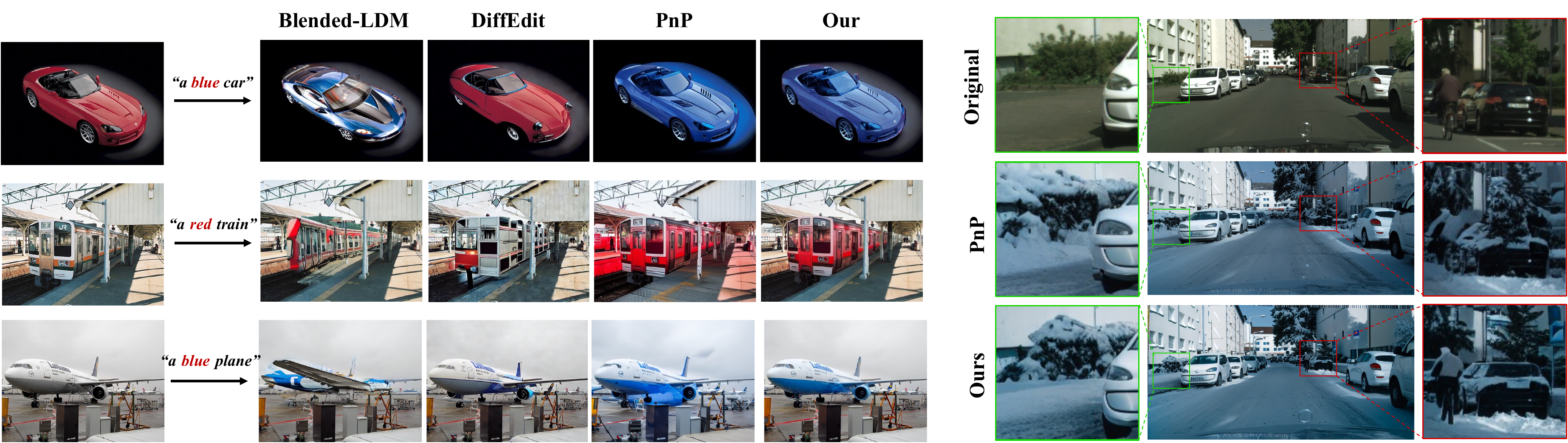}
    \caption{ Qualitative comparison of edited images. \textbf{Left:} Manipulating object color in Pascal VOC \cite{pascal}. \textbf{Right:} Changing to snowy day in Cityscapes \cite{cityscapes}. Our method achieves the best performance in structure preservation and object consistency.}
    \vspace{-1em}
    \label{fig:quality_assess}
\end{figure*}

The results are reported in Table \ref{table:image_assess}.
Firstly, previous approaches have the same performances of high scores in DINO Dist and LPIPS, but we argue that the reasons behind them are different.
Since DiffEdit \cite{diffedit} and Blended-LDM \cite{blended} use a mask to specify the edited region, they can ensure that irrelevant background information is not affected, but they do not achieve fine-grained control on object inner structures, \textit{e.g.} editing color will change other core properties.  PnP \cite{pnp} acts differently: it achieves precise control of the object's inner structural consistency but will disturb details of the adjacent background. Therefore, these approaches could impede the reliability of generated images for evaluation.
Secondly, we observe that since our method induces extra spatial constraints in the diffusion process, we achieve the best performance. As shown in Figure \ref{fig:quality_assess}, we can faithfully change object local attributes without affecting other information.
%Such observation is consistent to previous works \cite{sdedit, diffedit}. 
%To ensure the correctness of semantic layout labels after editing, and the reliability ity of the evaluation results provided by our framework, we have to make sacrifices in the realism of generated images. 
Finally, we also compare our methods with baseline in complex scenes, qualitative results are shown in Figure \ref{fig:quality_assess}. Our proposed method achieves improvement in semantic structure preservation under both object-centric images and complex scenes. 

In summary, we prove that existing editing approaches are not adequate as an evaluator for segmentation models, and demonstrate the reliability of our proposed pipeline. 

\begin{table}[t]
\centering
\caption{Quantitative results of different diffusion models in editing color and material attributes of objects on Pascal VOC \cite{pascal}.} 
\label{table:image_assess}
\resizebox{\columnwidth}{!}{
\begin{tabular}{l cc cc}
\toprule
\multirow{3}{*}{Method}  & \multicolumn{2}{c}{\textbf{Material}} & \multicolumn{2}{c}{\textbf{Color}} \\ 
\cmidrule(lr){2-3}  \cmidrule(lr){4-5} 
& DINO Dist ($\downarrow$)  &  LPIPS ($\downarrow$)   & DINO Dist ($\downarrow$)  &  LPIPS ($\downarrow$)  \\
\midrule
Blended-LDM \cite{blended}           &  0.081  & 0.397          &  0.079  & 0.375  \\
DiffEdit \cite{diffedit}             &  0.068  & 0.313          &  0.053  & 0.321  \\
Prompt-to-Prompt \cite{p2p}          &  0.044  & 0.367          &  0.030  & 0.472  \\
\midrule
PnP (Baseline) \cite{pnp}            & 0.052  & 0.319          &  0.047  & 0.408  \\
PnP w/ ControlNet \cite{controlnet}  & 0.010  & 0.284          &  0.010  & 0.365  \\
PnP w/ MGA                           & 0.017  & 0.250          &  0.012  & 0.309  \\
PnP w/ Both (Ours)                   & \textbf{0.003} & \textbf{0.185} &  \textbf{0.001}  & \textbf{0.156}    \\
\bottomrule
\end{tabular}}
\vspace{-1.3em}
\end{table}

\begin{table*}[]
\footnotesize
\caption{Comparison of our generated images with previous benchmarks under four adverse weather conditions, where generative and simulator refer to that images are generated by generative methods and simulator, respectively.}
%Our generated images are more realistic and reliable than existing synthetic benchmarks.
\vspace{-0.4em}
\label{table:benchmark_compare1}
\centering
\begin{tabular}{lc cc cc cc cc}
\toprule
 & \multirow{2}{*}{Model} & \multicolumn{2}{c}{Snow} & \multicolumn{2}{c}{Rain} & \multicolumn{2}{c}{Fog} & \multicolumn{2}{c}{Night} \\
 \cmidrule(lr){3-4} \cmidrule(lr){5-6} \cmidrule(lr){7-8} \cmidrule(lr){9-10}
 & & CLIP Acc $\uparrow$ & FID $\downarrow$  & CLIP Acc $\uparrow$ & FID $\downarrow$ & CLIP Acc $\uparrow$ & FID $\downarrow$ & CLIP Acc $\uparrow$ & FID $\downarrow$      \\
\midrule
ACDC \cite{acdc} & real &  1.000   &  0.000   &  1.000 & 0.000  & 1.000 &  0.000  &  1.000  &  0.000  \\
%Dark Zurich \cite{darkzurich} & Real &    - &     -     &    -      &    -    &    -      &  -    &   1.000    &          \\
\midrule
Multi-weather \cite{multiweather} &  \textit{generative} & 0.163 &  197.48  &  0.852   &  189.85 & - &  - &   0.906  & \textbf{154.30}   \\
Fog Cityscapes \cite{foggydriving} & simulator & - & - & -  & - &  0.940 & 164.00 & - &  -        \\
SHIFT \cite{shift} &  simulator & - &  - &  0.980   & 242.28 &  0.955  &  272.58  & 0.914 & 274.45  \\
Ours   & \textit{generative} &   \textbf{1.000}    &  \textbf{150.61}  & \textbf{0.989}  &  \textbf{143.47}  & \textbf{0.999}  & \textbf{137.15}  & \textbf{0.965}  &  186.69  \\
\bottomrule
\end{tabular}
\vspace{-0.4em}
\end{table*}

\begin{table}[]
\centering
\caption{Comparison of our pipeline with Stylized COCO \cite{trapped} in local object and background variations. Our generated images have superior image reality and fidelity.} 
\label{table:benchmark_compare2}
\resizebox{\columnwidth}{!}{
\begin{tabular}{lcccc}
\toprule
Benchmarks  & Model       & CLIP Acc $\uparrow$ & DINO Dist $\downarrow$  \\
\midrule
Stylized-object \cite{trapped}  & AdaIN \cite{adain} &   0.574  &  0.075   \\
Ours-object      & Diffusion &   \textbf{0.895}     &   \textbf{0.002}    \\
\midrule
Stylized-background \cite{trapped} & AdaIN \cite{adain} &  0.678  &   0.084   \\
Ours-background      & Diffusion &  \textbf{0.965}    &  \textbf{0.004}    \\
\bottomrule
\end{tabular}}
\vspace{-1em}
\end{table}

\noindent \textbf{Comparison with other benchmarks.}
Existing benchmarks \cite{acdc, shift, foggydriving, multiweather, foggydriving} for semantic segmentation robustness heavily focuses on adverse weather conditions of autonomous driving environments.
They utilize generative models \cite{adain, cyclegan} or simulator \cite{shift, foggydriving} to synthesize samples, to prevent manually collecting samples and labels. In this way, we argue that some properties of synthetic images could impede the reliability of evaluation results.  
We compare our edited images with previous benchmarks in terms of image reality, to demonstrate the reliability of our pipeline. We adopt (1) CLIP Acc \cite{clipscore}, which calculates the percentage of instances where the target image has a higher similarity to the target text than to the source text, (2) Fréchet Inception Distance (FID), which measures similarity between the distribution of real images and the distribution of generated images.

We separately generate images for four adverse kinds of weather (snow, rain, fog, and night) using the Cityscapes dataset \cite{cityscapes} as source images. And we adopt three existing synthetic benchmarks Multi-weather \cite{multiweather}, Fog Cityscapes \cite{foggydriving}, and SHIFT \cite{shift} as comparative methods. The FID score is computed by corresponding real images in ACDC \cite{acdc}.
The quantitative results are illustrated in Table \ref{table:benchmark_compare1}. 
In most weathers, our generated data are more realistic and distributional closer to real images than previous synthetic benchmarks. At night, our images are drastically greater than simulation-based benchmark SHIFT \cite{shift}, but worse than Multi-weather \cite{multiweather} which utilizes a set of GAN and CycleGAN models \cite{vqgan, cyclegan} to transfer styles. We think this is because diffusion models struggle in manipulating brightness and light information \cite{yin2023cle}.

Furthermore, since previous Stylized COCO \cite{trapped} generates local style variations on objects to study robustness in segmentation, we also compare our method with it. 
Following its setting, we generate images where objects and backgrounds have random art styles, respectively. From the results in Table \ref{table:benchmark_compare2}, our method significantly surpasses it in image reality and structural preservation. 
%Besides, we notice that previous stylized datasets own considerably greater FID scores, which implies that they could bring more out-of-distribution interference, and hence impede their reliability of evaluation.
In essence, we demonstrate that samples edited by our diffusion model have better image reality compared to conventional synthetic benchmarks for evaluating segmentation models. 
All experimental results can serve as evidence that our pipeline can be a better substitution for real images, especially since acquiring real samples and labels is laborious.

\section{Conclusion}
In this paper, we provide a pipeline that can precisely edit the visual attributes of real images and preserve their original mask labels.
With this pipeline, we construct a benchmark and evaluate the robustness of diverse segmentation models against different object and image attribute variations.
Experimental results reveal that most models are vulnerable to object attribute changes. Meanwhile, advanced models with stronger backbones and massive training data do not necessarily show better robustness. 
Our experiments suggest object attributes should be taken into account for improving segmentation robustness.
We also demonstrate the quality of our edited images and their reliability as test samples by comparing them with popular synthetic benchmarks and existing image editing methods.

%While we take a step forward to segmentation robustness evaluation, 
Our work has some limitations. First, due to the failure modes of the diffusion model, it is difficult to edit the attributes of the person. Second, since inherent biases in diffusion models, altering an attribute could occasionally change other appearance information. For example, sometimes editing an object's material to wood may deviate its original color to brown. We would like to explore how to avoid the spurious editing problem in future work.
\vskip0.2cm
\noindent \textbf{Acknowledgements} 
The authors thank Xinran Wang for his valuable feedback and discussion. 
This work was supported in part by National Natural Science Foundation of China (NSFC) No. 62106022, 62225601, U23B2052, in part by Beijing Natural Science Foundation Project No. Z200002 and in part by Youth Innovative Research Team of BUPT No. 2023QNTD02.

{
    \small
    \bibliographystyle{ieeenat_fullname}
    \bibliography{main}
}


\end{document}

\maketitle
%\maketitlesupplementary
%\appendix
\let\thefootnote\relax\footnotetext{$\ast$ Corresponding author.}

%\tableofcontents

\section{More details on Mask-Preserved Attribute Editing Pipeline }
%\subsection{Caption Generation Details}
%We provide the details we generate image captions from Pascal VOC \cite{pascal} using BLIP2 \cite{blip2}. The 

\subsection{Text Manipulation details}
In Table \ref{table:prompts}, we introduce exhaustive prompts used to instruct GPT-3.5 Turbo \cite{llms}, and edited sentence examples with different attribute variations. We find that adding pre-defined roles (\textit{the professional linguistic assistant}) and examples in text prompts can drastically improve performances.

\begin{table*}[]
\centering
\caption{Illustrations of text prompts used to instruct GPT-3.5 Turbo \cite{llms} and edited sentence examples. Edited parts are colored red.}
\label{table:prompts}
\resizebox{\textwidth}{!}{
\begin{tabular}{c|p{12cm}|p{6cm}}
\toprule
Variation type & Text Prompt & Examples \\
\midrule
Local color          &  \textit{You are a professional linguistic assistant. I will provide you with a sentence. You should first identify the subject of the sentence, and then generate a variation by altering or adding the color attributes of the subject. You should only return the sentence variation. For example, change "a photo of an airplane on the ground" to "a photo of a blue airplane on the ground"}      & Input: \textit{a photo of a white and red train.} Output: \textit{a photo of a \textcolor{red}{blue and yellow} train.}        \\
\midrule
Local material       &  \textit{You are a professional linguistic assistant. I will provide you with a sentence. You should first identify the subject of the sentence, and then generate a variation by changing the material attribute of the subject. The material should be selected from "wooden", "paper", "metallic" and "paper". You should only return the sentence variation. For example, change "a photo of an airplane on the ground" to "a photo of a wooden airplane on the ground"}  & Input: \textit{a photo of a white and red train.} Output: \textit{a photo of a \textcolor{red}{wooden} white and red train.}       \\
\midrule
Local pattern        &  \textit{You are a professional linguistic assistant. I will provide you with a sentence. You should first identify the subject of the sentence, and then generate a variation by changing the material attribute of the subject. The type of patterns should be selected from "dotted", "striped" and "lettered". You should only return the sentence variation. For example, change "a photo of an airplane on the ground" to "a photo of an airplane with stripes on the surface on the ground".}            &  Input: \textit{a photo of a white and red train.} Output: \textit{a photo of a \textcolor{red}{striped} white and red train.}      \\
\midrule
Global domain:      & \textit{You are a professional linguistic assistant. You should generate one possible edition by changing the provided sentence's data domain without changing the content. The data domain should be selected from "oil pastel", "painting", "and sketch" For example, change "a photo of an airplane on the ground" to "a painting of an airplane on the ground"}  &  Input: \textit{a photo of a white and red train.} Output: \textit{a \textcolor{red}{sketch} of a white and red train.}    \\
\midrule
Global weather:     & \textit{You are a professional linguistic assistant. You should generate one possible edition by only adding a weather description to the provided sentence without changing the content. The weather should be selected from "snow", "rain", and "fog". For example: change "a photo of an airplane on the ground" to "a photo of an airplane on the ground on a snowy day".}   & Input: \textit{a photo of a white and red train.} Output: \textit{a photo of a red and white train \textcolor{red}{against a backdrop of falling snow}}        \\
\bottomrule
\end{tabular}}
\end{table*}

\subsection{Mask-Guided Diffusion details}
\begin{table}[]
\footnotesize
\centering
\caption{Parameter settings of Mask-Guided Diffusion.}
\label{table:parameters}
\begin{tabular}{c|c}
\toprule
Parameters & Values \\
\midrule
Image resolution         & 512$\times$512       \\
SD version               & 1.5       \\
Seed                     & 1         \\
Guidance scale           & 7.5       \\
Inversion timesteps      & 1000      \\
Diffusion timesteps      & 50       \\
$\tau_f$ \cite{pnp}                & 0.8       \\
$\tau_A$ \cite{pnp}                & 0.5       \\
$\tau_c$ (ours)                & 0.5       \\
$\tau_m$ (ours)               & 0.0       \\
\bottomrule
\end{tabular}
\end{table}

We leverage the state-of-the-art text-to-image algorithm Latent Diffusion Model \cite{ldm}, a.k.a Stable Diffusion (SD), in which the diffusion process performs in low-dimensional latent space where semantic information can better transfer. It consists of a variational autoencoder network to encode and decode between latent space and pixel space, and denoising network U-Net \cite{unet} architecture conditioned on the guiding text prompt to achieve diffusion process. And we integrate our Mask-Guided Attention and ControlNet \cite{controlnet} to the text-guided Image-to-Image Translation approach PnP \cite{pnp}. 
In all our results, the Mask-Guided Attention and ControlNet block \cite{controlnet} is integrated into all decoder layers of Stable Diffusion. 
For integration duration in the denoising process, we utilize two thresholds (i) $\tau_m \in [0,1]$ is the sample step until Mask-Guided Attention is integrated, (ii) $\tau_c \in [0,1]$ is the sample step until which ControlNet block \cite{controlnet} is integrated. We set $\tau_m = 0$ since we need to ensure the irrelevant region is not affected in every step of the denoising process. We set $\tau_c = 0.5$ since a large value will diminish the spatial constraint effects of semantic layout labels, and a small value will reduce the reality of edited images. More discussion on $\tau_c$ is presented in Sec \ref{subsec:tau_c}.
The detailed parameter values in our pipeline are presented in Table \ref{table:parameters}.

It is noteworthy that the Mask-Guided Attention serves in local attribute editing and does not serve in global attribute editing. This is because in local editing we need to identify the area of edition by the object mask, while in global editing the whole image needs to be changed.

%We introduce more details of Mask-Guided Diffusion which integrates our Mask-Guided Attention and ControlNet \cite{controlnet} to image editing approach PnP \cite{pnp}. The baseline PnP \cite{pnp} first computes the initial noise of the original image by DDIM \cite{ddim}, and then obtains spatial features and self-attention features from the denoising process of the original image and directly injects them into the denoising process of edited images.  

\section{More Evaluation Results}

\subsection{Per-Class analysis}

\begin{figure*}
    \centering
    \includegraphics[width=\textwidth]{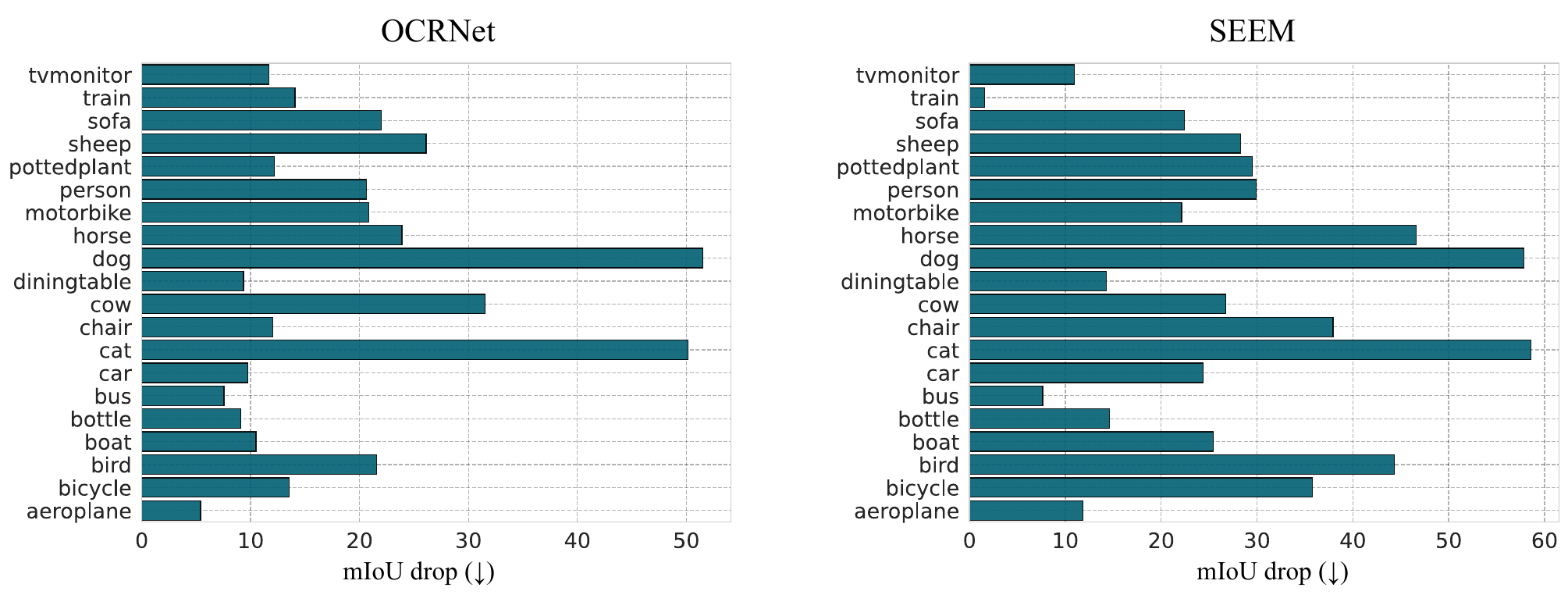}
    \caption{Average mIoU drop ($\uparrow$) in each class of OCRNet \cite{ocr} and SEEM \cite{seem} under our Pascal-EA.}
    \label{fig:perclass}
\end{figure*}

Figure \ref{fig:perclass} shows the per-class mIoU drop of two segmentation models OCRNet \cite{ocr} and SEEM \cite{seem}. It can be observed that creature classes like dog, cat, and horse are more easily disturbed than inanimate object classes such as bus and train.

\subsection{Robustness comparison}
\begin{figure}
    \centering
    \includegraphics[width=\linewidth]{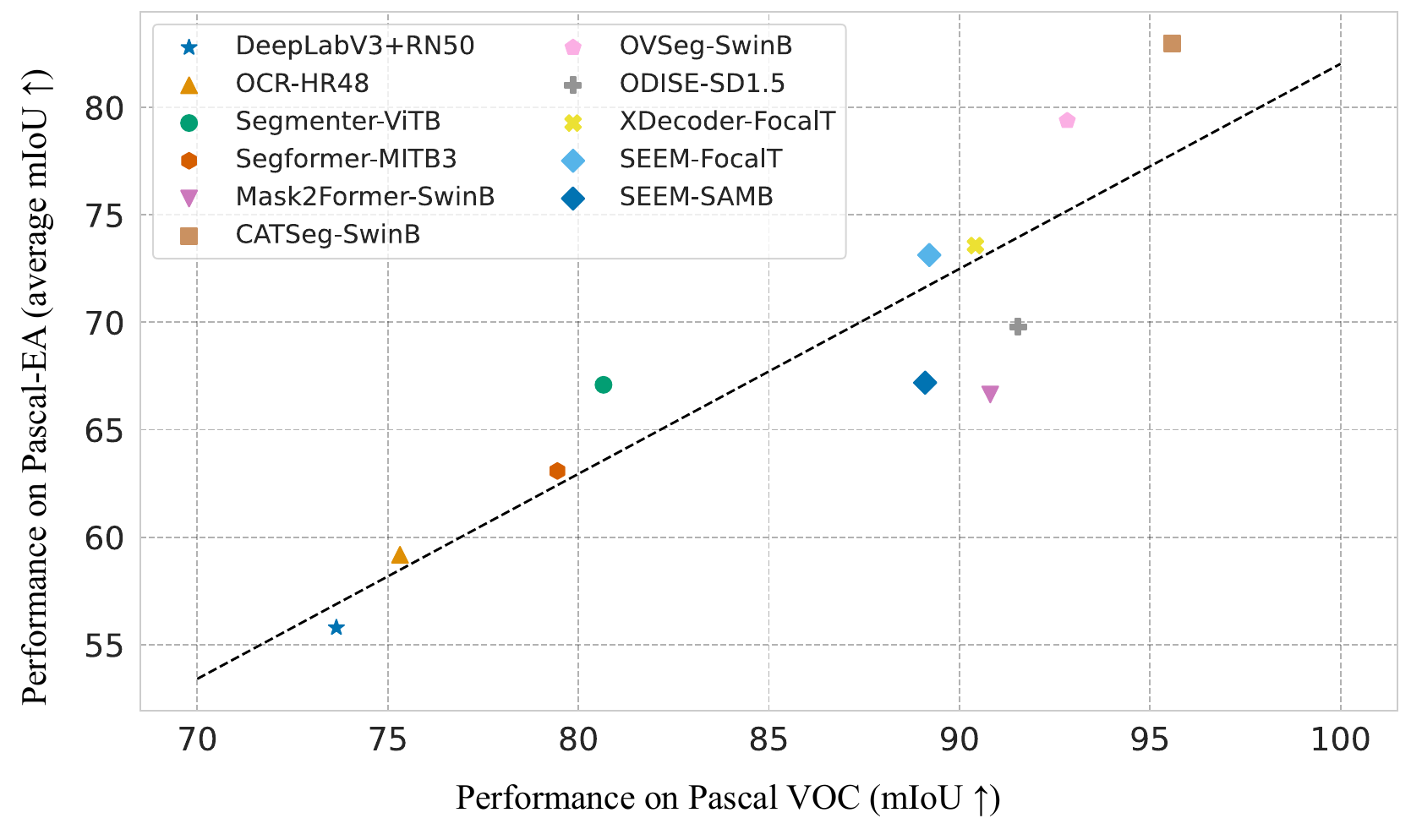}
    \caption{The average performances on our Pascal-EA \textit{vs.} Performances on original Pascal VOC \cite{pascal}. The black dashed line indicates the linear fit of all segmentation methods.}
    \label{fig:robustness}
\end{figure}

To exclude the impact of the performance improvement in original data on our benchmark, we evaluate model robustness by calculating the segmentation accuracy decline in Section 4. In this part, we additionally plot the mIoU accuracy on original Pascal VOC \cite{pascal} \textit{vs.} our Pascal-EA benchmark. As shown in Figure \ref{fig:robustness}, the CATSeg \cite{cat} exhibits the greatest robustness than others.

\subsection{More qualitative results}
\begin{figure*}
    \centering
    \includegraphics[width=\textwidth]{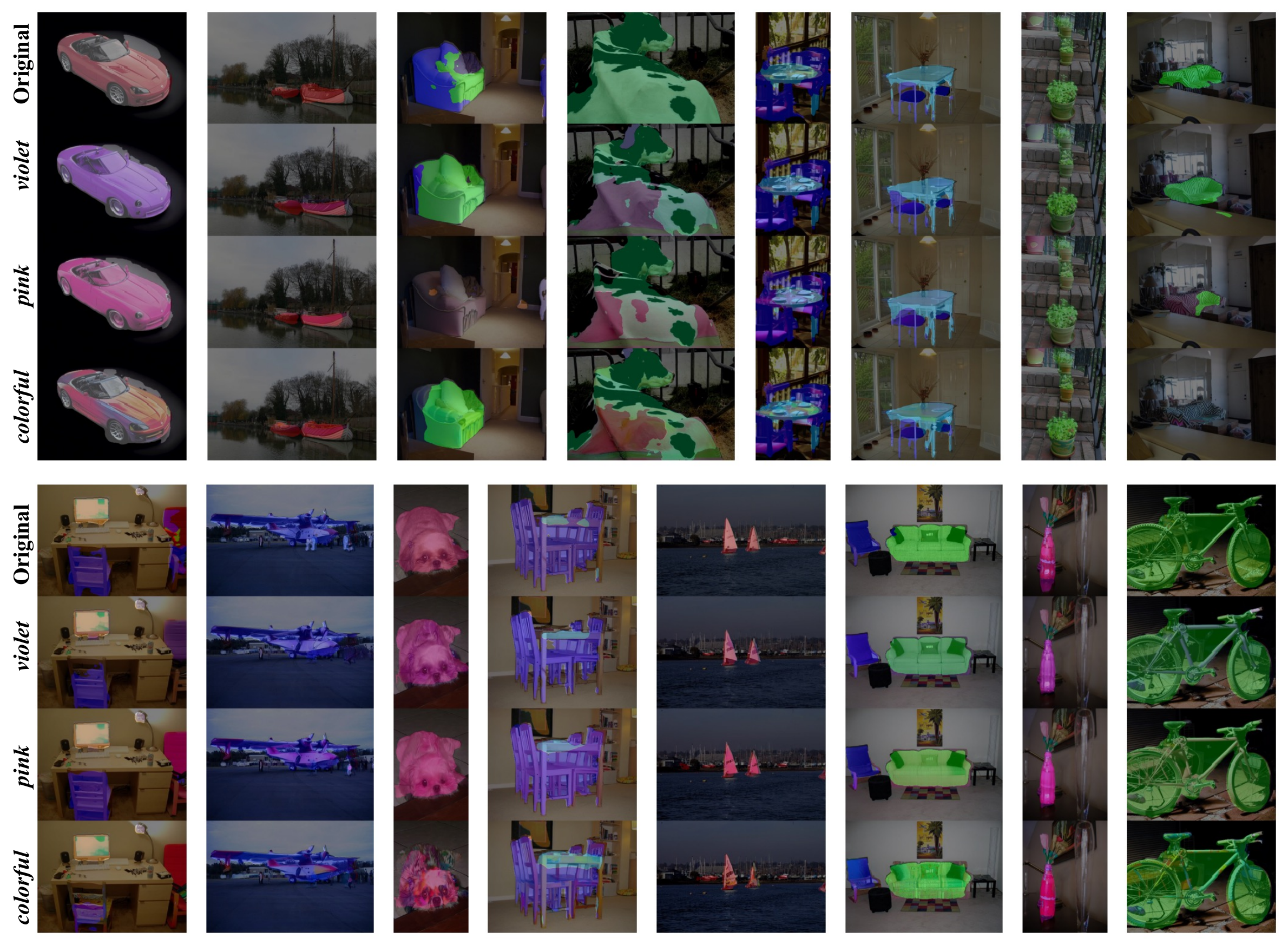}
    \caption{Qualitative segmentation results of OCRNet \cite{ocr} under object color attribute variations.}
    \label{fig:seg_visualization_color}
\end{figure*}

\begin{figure*}
    \centering
    \includegraphics[width=\textwidth]{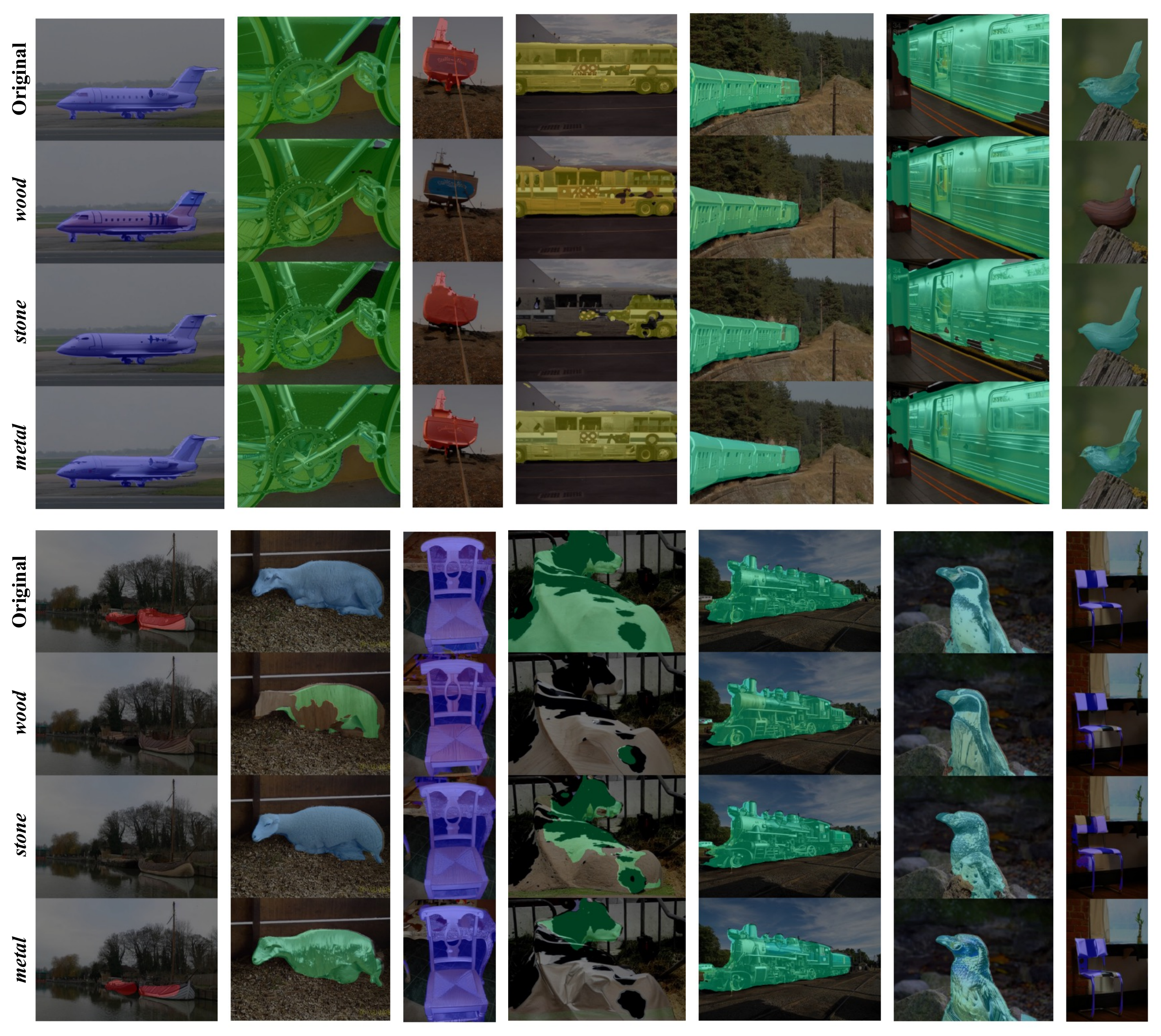}
    \caption{Qualitative results of OCRNet \cite{ocr} under object material attribute variations.}
    \label{fig:seg_visualization_material}
\end{figure*}

\begin{figure*}
    \centering
    \includegraphics[width=\textwidth]{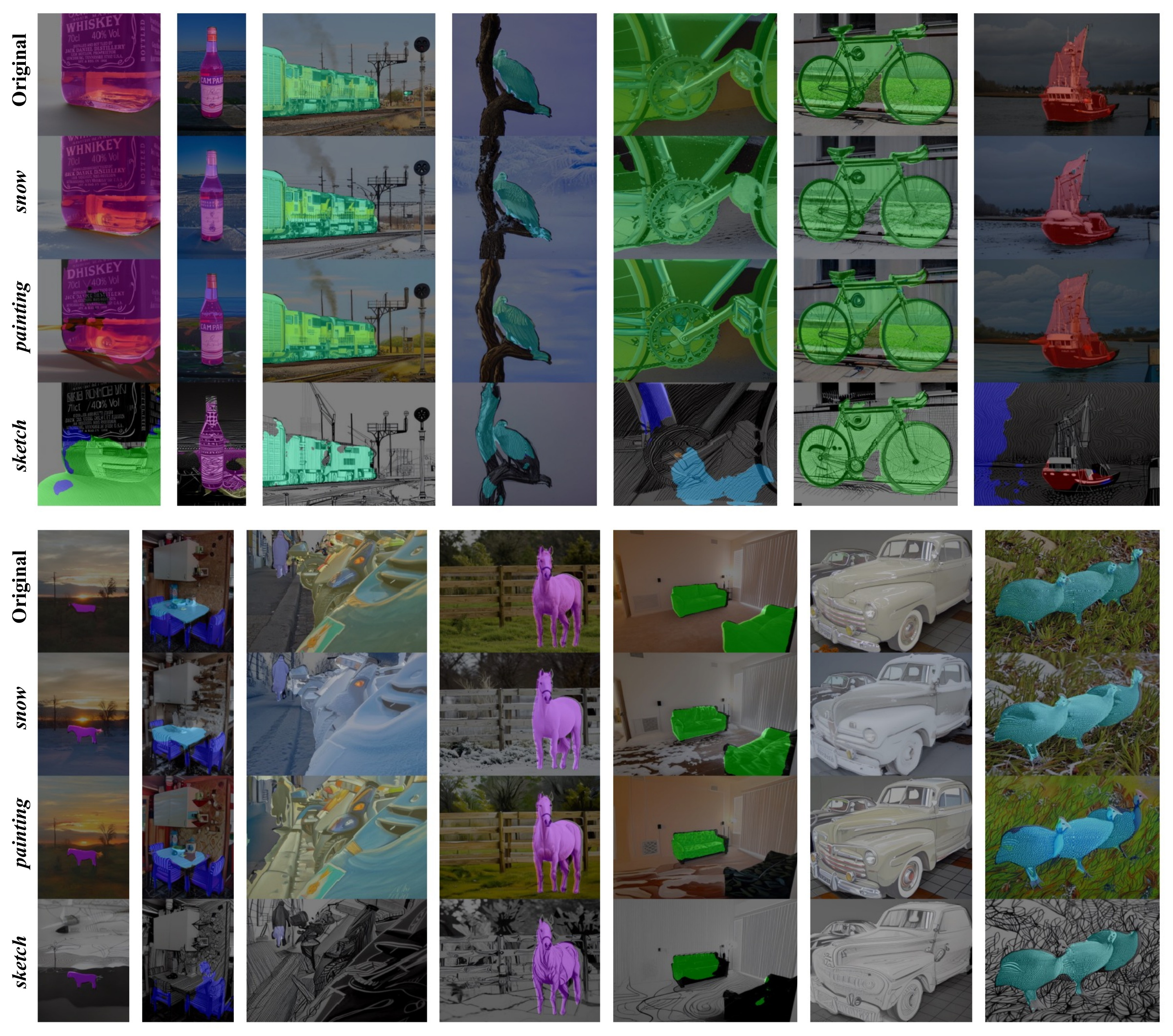}
    \caption{Qualitative segmentation results of OCRNet \cite{ocr} under image style attribute variations.}
    \label{fig:seg_visualization_style}
\end{figure*}

Figures \ref{fig:seg_visualization_color}, \ref{fig:seg_visualization_material} and \ref{fig:seg_visualization_style}  illustrate qualitative results of segmentation methods under different attribute variations. 

\section{Analysis of Pascal-EA}
\begin{figure*}
    \centering
    \includegraphics[width=\textwidth]{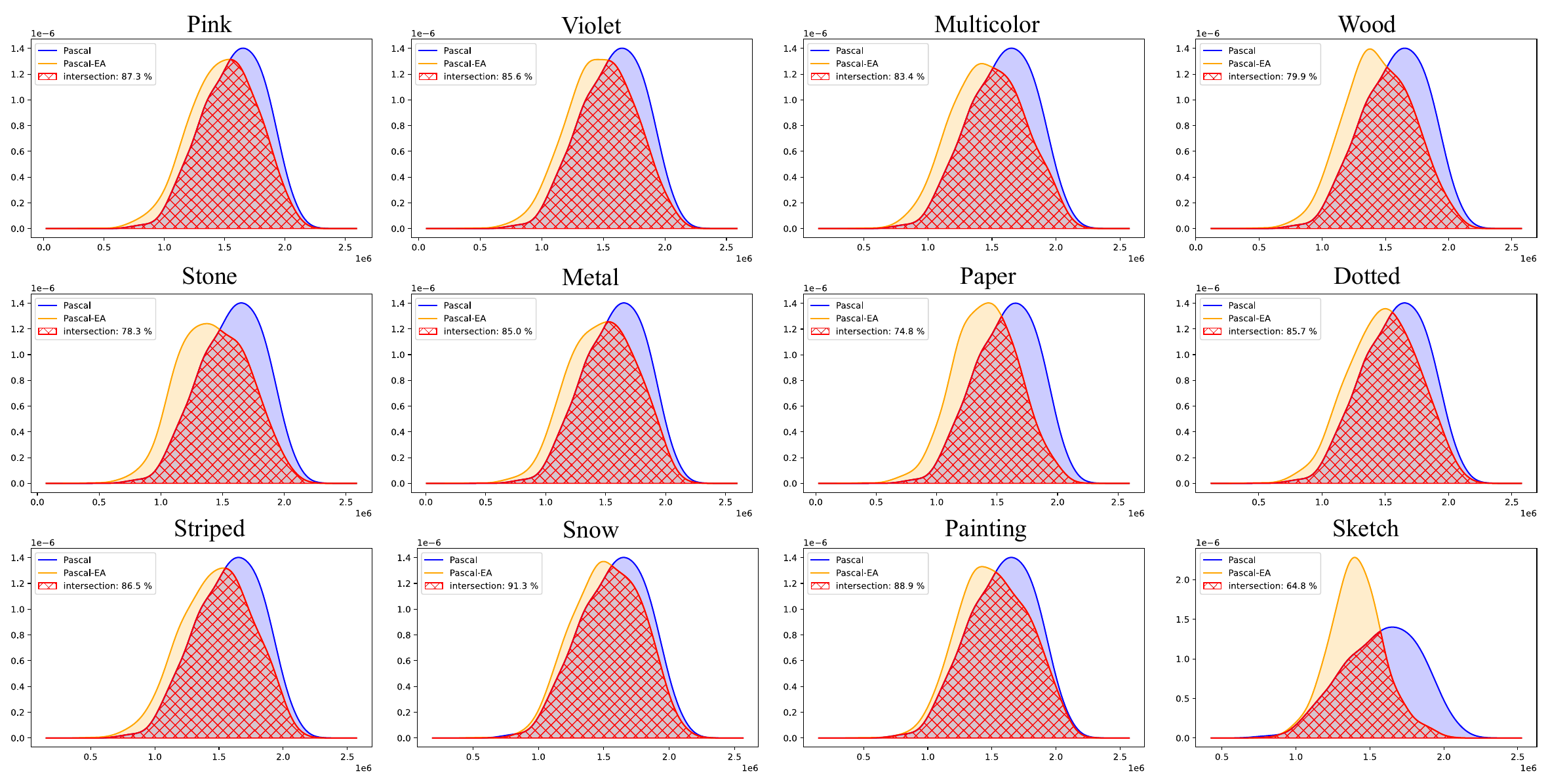}
    \caption{Different distribution of our edited images and original Pascal VOC \cite{pascal} in terms of the quantities in GradNorm.}
    \label{fig:gradnorm}
\end{figure*}
In contrast to previous benchmarks like COCO-O \cite{coco-o} and ImageNet-C \cite{weather} providing samples from out-of-distribution domains, our Pascal-EA consists of image variations in editable attributes within the in-distribution domain. We measure the extent to which the distribution of our Pascal-EA shifts from that of the original Pascal VOC dataset \cite{pascal} by the widely-used out-of-distribution detection approach GradNorm \cite{gradnorm}. Experimental results in twelve different attribute variations are shown in Figure \ref{fig:gradnorm}, the x-axis is gradient norms scores and the y-axis is the density of each score, and we calculate the overlap area of two distributions in which heavier overlap indicates closer to original distribution. We observe that our Pascal-EA can proximity the original distribution meanwhile having object attribute variations, which implies can provide reliable evaluations.

To explore whether the diffusion process will degrade the reliability of our evaluations for segmentation, we create a reference set in which we reconstruct the original validation set using our pipeline with non-edited texts. This operation first adds noises to the original images and then denoise them with non-edited texts.
We compare model performances on the original validation set and our reference set in Table \ref{table:reference}. It is obvious that the diffusion process induces subtle disturbing on segmentation performances which is negligible compared to attribute variation themselves. 
Such results also serve as evidence that our pipeline’s robustness to potential errors in generated caption.
In our all experiments in Sec 4, we replace results in the validation set with those in our reconstructed set to remove the effects of perturbations.

\begin{figure*}
    \centering
    \includegraphics[width=\textwidth]{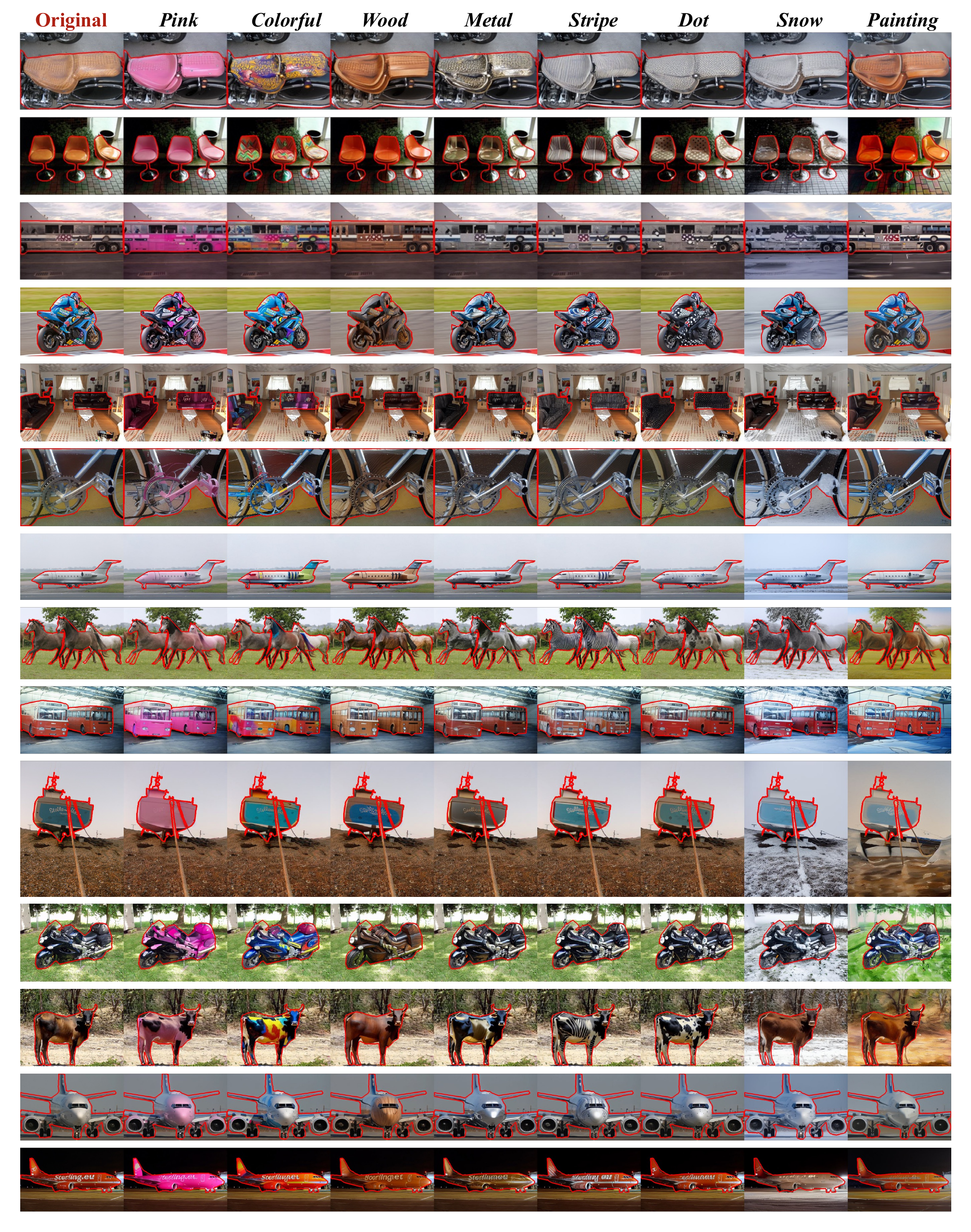}
    \vspace{-2em}
    \caption{Resulting images edited by our method. Our method effectively converts various objects into versions with  different attributes,  while ensuring their segmentation mask is consistent with the original ones.}
    \label{fig:edition_visualization}
\end{figure*}

\begin{figure*}
    \centering
    \includegraphics[width=\textwidth]{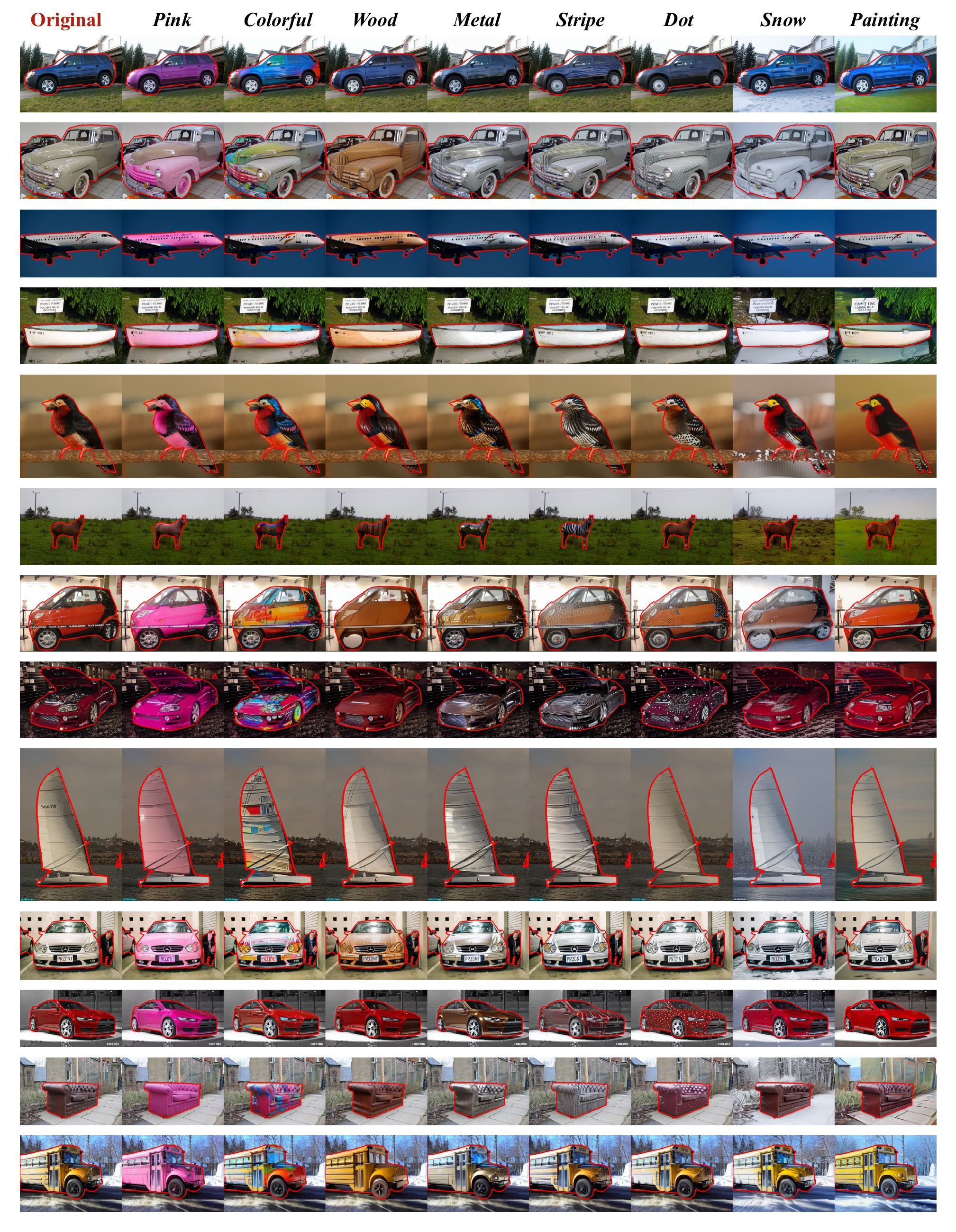}
    \vspace{-25pt}
    \caption{Resulting images edited by our method. Our method effectively converts various objects into versions with  different attributes,  while ensuring their segmentation mask is consistent with the original ones.}
    \label{fig:edition_visualization2}
\end{figure*}

Additional qualitative results of the Pascal-EA benchmark are illustrated in Figure \ref{fig:edition_visualization} and Figure \ref{fig:edition_visualization2}. The red lines delineate the boundaries of objects in ground truths, our mask-preserved pipeline can ensure the correctness of original labels in attribute-edited images.

\begin{table*}[]
\centering
\footnotesize
    \begin{tabular}{l|c|ccccc|c|ccccc}
    \toprule
         & \multicolumn{6}{c|}{HRNet \cite{hrnet}} & \multicolumn{6}{c}{Segformer \cite{segformer}} \\ 
        Method & CS & Rain & Fog & Snow & Night & Avg. & CS & Rain & Fog & Snow & Night & Avg.\\ 
        \midrule
        Baseline  & 70.47 & 44.15 & 58.68 & 44.20 & 18.90 & 41.48  & 67.90 & 50.22 & 60.52 & 48.86 & 28.56 & 47.04\\ 
        \midrule
        CutOut \cite{cutout}   & 71.39 & 40.29 & 57.70 & 43.98 & 16.55 & 39.63  & 68.93 & 47.68 & 60.34 & 46.98 & 26.49 & 45.37     \\
        CutMix \cite{cutmix}   & 72.68 & {42.48} & {58.63} & 44.50 & {17.07} & {40.67} & {69.23} & {49.53} & 61.58 & {47.42} & {27.77} & {46.57} \\
        Weather \cite{weather}   & 69.25 & \textbf{50.78} & 60.82 & {38.34} & 22.82 & 43.19  & 67.41 & \textbf{54.02} & 64.74 & 49.57 & 28.50 & 49.21 \\
        StyleMix \cite{stylemix} & 57.40 & {40.59} & {49.11}  & {39.14} & 19.34 & {37.04} & 65.30 & 53.54 & 63.86 & 49.98 & 28.93 & 49.08\\ 
        \textbf{Ours}    & 65.77 & 46.40 & \textbf{61.61} & \textbf{49.78} & \textbf{28.49} & \textbf{46.57} & 63.48 & 52.25 & \textbf{69.54} & \textbf{56.20} & \textbf{30.12} & \textbf{52.03}   \\
        \midrule
        Oracle & - & 65.67 & 75.22 & 72.34 & 50.39 & 65.90  & - & 63.67 & 74.10 & 67.97 & 48.79 & 63.56 \\
    \bottomrule
    \end{tabular}
\caption{Comparison of different methods from Cityscapes (source) to ACDC (target) using the mIoU ($\uparrow$) metric. The results are reported on the Cityscapes (CS) validation set, four individual scenarios of ACDC, and the average (Avg). The best performances are bold. Oracle indicates the supervised training on ACDC, serving as an upper bound for the other methods.}
\label{table:application1}
\end{table*}

\begin{table}[]
\centering
\resizebox{\linewidth}{!}{
\begin{tabular}{ccccccc}
\toprule
Vanilla & CutOut \cite{cutout} & CutMix \cite{cutmix} & \makecell{Digital \\ Corruption \cite{weather}}  & AugMix \cite{augmix} & \textbf{Our}\\
\hline
90.81 & 91.44 & 92.01 & 91.19 & 91.88 & \textbf{92.57}\\
\bottomrule
\end{tabular}}
\caption{The results of different data augmentation techniques using Mask2Former \cite{mask2former} on Pascal VOC dataset \cite{pascal}.}
\label{table:application2}
\end{table}

\begin{table}[]
\footnotesize
\centering
\caption{ The mIoU ($\uparrow$) of different methods on Pascal VOC \cite{pascal} and our Pascal-EA. The performance of all methods drops on  our Pascal-EA. }
\label{table:reference}
\begin{tabular}{lcc}
\toprule
Method & Pascal VOC & Pascal-EA \\
 \midrule
DeepLabV3+ \cite{deeplabv3plus} &  75.33          &  73.65 (\textcolor{red}{-1.68})         \\
OCRNet \cite{ocr} & 76.92           &  75.32 (\textcolor{red}{-1.60})         \\
Segmenter \cite{segmenter} &  82.25          &  80.66 (\textcolor{red}{-1.59})        \\
Segformer \cite{segformer} &  81.01          &  79.45 (\textcolor{red}{-1.56})        \\
Mask2Former \cite{mask2former} & 91.98       &  90.81 (\textcolor{red}{-1.17})        \\
\midrule
CATSeg \cite{cat} &  96.60          &  95.59 (\textcolor{red}{-1.1})        \\
OVSeg \cite{ovseg} & 94.49           &  92.83  (\textcolor{red}{-1.66})       \\
ODISE \cite{odise} & 93.22           &  91.54  (\textcolor{red}{-1.68})       \\
X-Decoder \cite{x-decoder} & 91.77       &  90.42 (\textcolor{red}{-1.35})        \\
SEEM \cite{seem} & 91.06           &  89.21 (\textcolor{red}{-1.85})        \\
\bottomrule
\end{tabular}
\end{table}

\section{Additional applications of our pipeline}

\textbf{Improving the generalization ability of segmentation methods.} 
We show that our mask-preserved attribute editing pipeline can be exploited to generate training images for improving the robustness ability of segmentation models. 
We construct edited training sets with four adverse conditions (fog, snow, rain, and night) using the Cityscapes dataset \cite{cityscapes}, and use them to train models.
Following the previous setting of \cite{issa}, we report performances from Cityscapes to ACDC domain generalization, the results of comparative approaches are directly from \cite{issa}. 
The quantitative results in Table \ref{table:application1} and \ref{table:application2} exhibit that model training with our data has competitive performances, and consistently gains improvement across all datasets and scenarios.
In Table \ref{table:application1}, since CutOut \cite{cutout} and CutMix \cite{cutmix} just combine local visual contents, it exhibits improvement in in-distribution performance while deterioration on global style shifts. On account of unreal images, Hendrycks-Weather \cite{weather} degrade performance in snow, and StyleMix \cite{stylemix} has declined in all scenarios.

\section{More discussion}
\begin{table}[]
\scriptsize
\centering
\caption{The quantitative assessments of text variations to original ones in Pascal VOC dataset \cite{pascal}.}
\label{table:text_assess}
\resizebox{\linewidth}{!}{
\begin{tabular}{ccccc}
\toprule
\multirow{2}{*}{}
& \multirow{2}{*}{Perplexity ($\downarrow$)} 
& \multirow{2}{*}{CIDEr ($\uparrow$)} 
& \multirow{2}{*}{SPICE ($\uparrow$)} 
& \multirow{2}{*}{\shortstack[c]{BERT-\\ Score ($\uparrow$)}} \\ \\
\midrule      
Source  & 103.71   & 10.00 & 1.00  &  0.71  \\
\midrule
\multicolumn{5}{l}{\textit{Text variations}} \\
\midrule
Color     & 107.70   & 7.34  & 0.85  &  0.71  \\
Material     & 118.14   & 3.58  & 0.56  & 0.65   \\
Pattern     & 124.26   & 6.10  & 0.71  &  0.62  \\
Style     & 129.54   & 7.35  & 0.67  &  0.69  \\
%\midrule
%Source  & 734.26     & 10.00 & 1.00  &  1.00  \\
%Target     & 238.72     & 1.49  & 0.61  &  0.85  \\
\bottomrule
\end{tabular}}
\end{table}

\subsection{Text quality evaluation} 

As generating variations of text descriptions of images by LLM \cite{llms}, our framework inevitably imports text perturbations. 
To extensively evaluate the quality of target texts in comparison to the source, we adopt several metrics: (1) Perplexity to measure sentence quality, (2) CIDEr and SPICE to measure the fidelity and semantic meanings respectively, (3) BERT score \cite{bert} which calculate cosine similarity between texts and category labels of images to measure the consistency with class ground truth. The results are shown in Table.\ref{table:text_assess}. The results indicate a reduction in both the quality and semantic consistency of generated texts, but the decrement is in a reasonable range. Thus, we infer that while the LLM introduces additional noise to texts, it is still adequate as a text editor in our framework.

\subsection{Impact of threshold $\tau_c$}
\label{subsec:tau_c}

\begin{figure*}
    \centering
    \includegraphics[width=\textwidth]{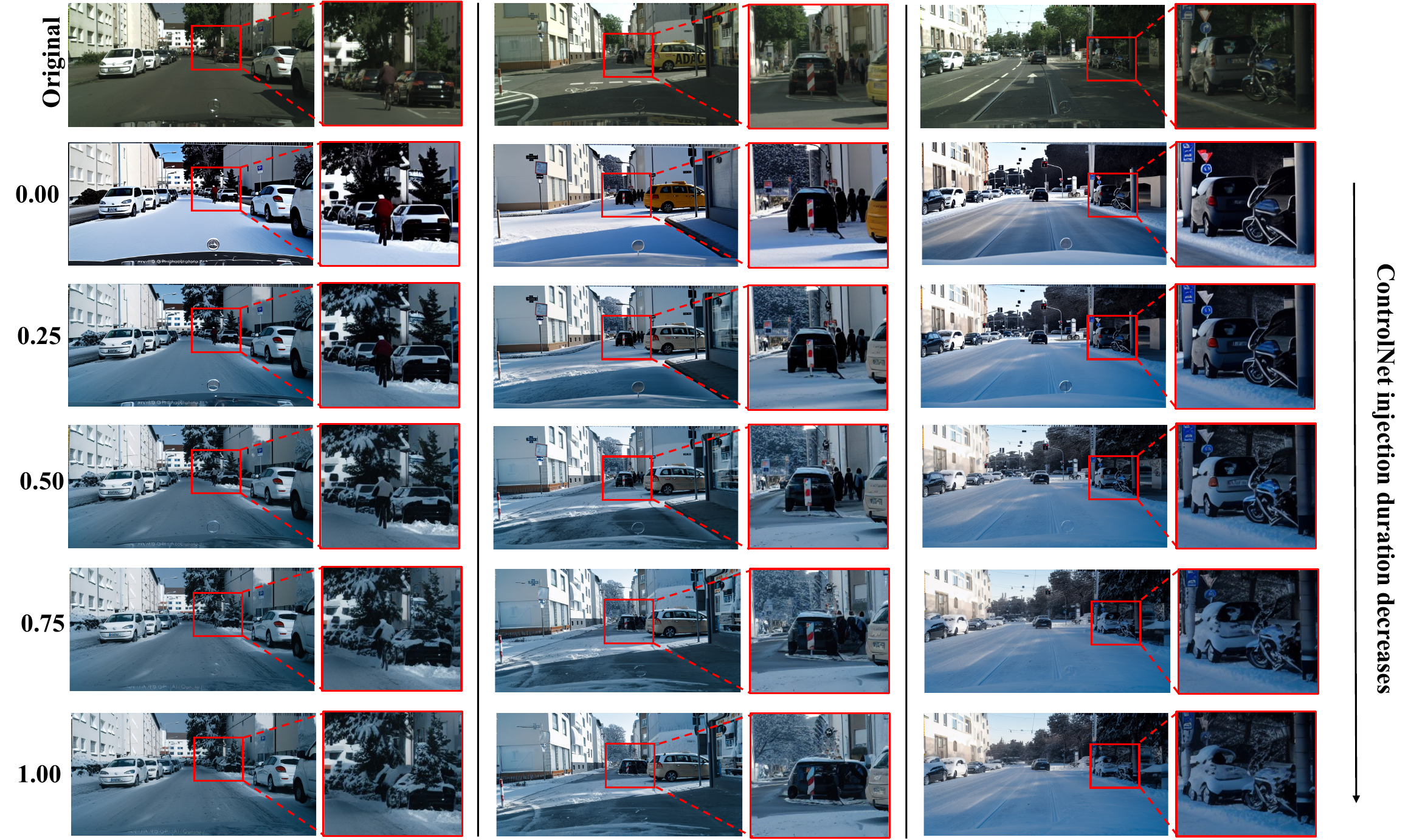}
    \caption{Resulting images edited by our method using different values of $\tau_c$}
    \label{fig:controlnet}
\end{figure*}

\begin{figure}
    \centering
    \includegraphics[width=\linewidth]{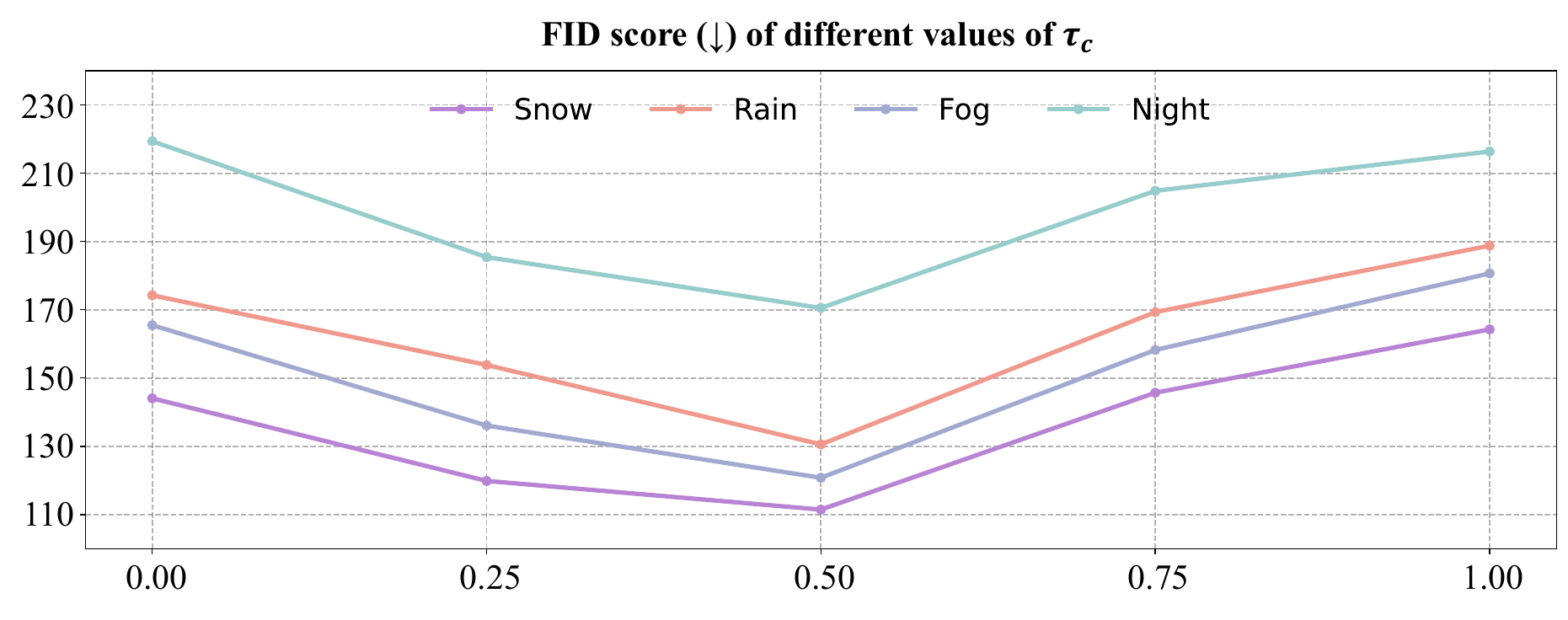}
    \caption{FID score ($\downarrow$) of our generation results using different values of $\tau_c$. When $\tau_c=0.5$,  our method achieves the best performances in all weather conditions.}
    \label{fig:controlnet_fid}
\end{figure}
The threshold $\tau_c \in [0,1]$ defines the duration of ControlNet \cite{controlnet} injection during the denoising process, the smaller value indicates a longer adoption duration. 
We perform additional experiments to explore its effects on edited image quality. 
We translate images with complex city scenes in the Cityscapes dataset \cite{cityscapes} to that on adverse weather (snow, rain, fog, night). Figure \ref{fig:controlnet} illustrates several qualitative results. We observe that, as ControlNet \cite{controlnet} injection duration decreases, the structural consistency decreases while the realism of edited images increases. 
To make a trade-off between realism and structural consistency, we further compute their FID score as shown in Figure \ref{fig:controlnet_fid}. We can notice that as $\tau_c=0.5$ we achieve the best performances on all weather editing scenarios. Therefore, we adopt $\tau_c=0.5$ in our pipeline.

%\subsection{Mask-Guided Attention and ControlNet}
%In Sec \ref{subsec:compare_diffusion}, we demonstrate the effectiveness of Mask-Guided Attention and ControlNet in spatial constraint.
%To further explore their differences, we conduct additional ablation studies in both local and global image editing scenarios and report qualitative results in Figure \ref{table:module_ablation}. We need to clarify that since our Mask-Guided Attention serves in local editing, corresponding results of PnP w/ Mask-Guided Attention in global editing scenarios are ignored. 
%In summary, the effectiveness of Mask-Guided Attention and ControlNet differ in local and global editing scenarios, since ControlNet explicitly restricts the image semantic layout, and Mask-Guided Attention designates local edited area. 

\subsection{Failure cases of edited images}
\begin{figure}
    \centering
    \includegraphics[width=\linewidth]{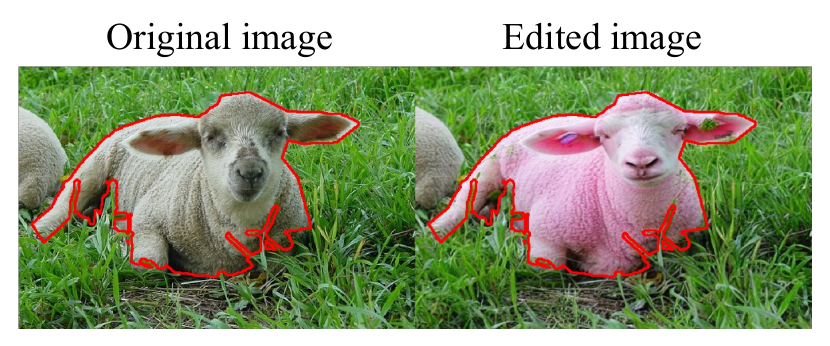}
    \caption{Illustration of our failure cases.  Our method does not faithfully preserve the appearance of the face due to the limitations of the diffusion model; however, our method effectively preserves the global structure, which ensures the ground-truth mask of the edited image is consistent with that of the original one.}
    \label{fig:failure}
\end{figure}

Inheriting the innate limitations of diffusion models, our pipeline has unpleasant performances in several scenarios. The qualitative failure cases of edited images are shown in Figure \ref{fig:failure}. As multiple objects overlap, our edited images violate the original structures of inside objects. 
Moreover, since the inherent failure modes in diffusion model \cite{ldm}, our edited images could deviate from the original face of creatures.

%\clearpage

%\clearpage

%\clearpage

%\clearpage

\clearpage
\clearpage

{
    \small
    \bibliographystyle{ieeenat_fullname}
    \bibliography{supplementary}
}